\documentclass[letterpaper, 10 pt, conference]{ieeeconf}  

\IEEEoverridecommandlockouts                              

\usepackage{epsfig} 
\usepackage{amsmath} 
\usepackage{amssymb}  
\usepackage{url}
\usepackage{balance}
\usepackage{caption}
\usepackage{gensymb}
\usepackage{subfig}

\usepackage{graphicx}
\usepackage{amsmath}
\usepackage{amssymb}
\usepackage{booktabs}
\usepackage{algorithm}
\usepackage{algorithmic}
\usepackage{array}
\usepackage{color, colortbl}
\usepackage{hyperref}

\newtheorem{definition}{Definition}

\definecolor{ashgray}{rgb}{0.7,0.75,0.71}
\definecolor{babypink}{rgb}{0.96,0.76,0.76}
\definecolor{taupegray}{rgb}{0.55, 0.52, 0.54}
\definecolor{gainsboro}{rgb}{0.86, 0.86, 0.86}
\definecolor{lightgray}{rgb}{0.83, 0.83, 0.83}
\definecolor{indigo(dye)}{rgb}{0.0, 0.25, 0.42}
\definecolor{silver}{rgb}{0.75, 0.75, 0.75}

\title{\LARGE \bf
RAPid-Learn: A Framework for Learning to Recover for \\Handling Novelties in Open-World Environments.
}

\author{
    \authorblockN{Shivam Goel*$^{1}$, Yash Shukla*$^{1}$, Vasanth Sarathy$^{2}$, Matthias Scheutz$^{1}$ and Jivko Sinapov$^{1}$}
    \thanks{\textit{(*equal contribution)}}%
    \thanks{$^{1}$ Department of Computer Science, Tufts University, Email: {\tt\small  \{shivam.goel, yash.shukla, matthias.scheutz, jivko.sinapov\} @ tufts.edu}}%
    \thanks{$^{2}$ Smart Information Technology (SIFT), Email: {\tt\small vsarathy@sift.net}}%
}

\begin{document}

\maketitle
\thispagestyle{empty}
\pagestyle{empty}

\begin{abstract}

    We propose \textit{RAPid-Learn} (\textit{Learn}ing to \textit{R}ecover and \textit{P}lan \textit{A}gain), a hybrid planning and learning method, to tackle the problem of adapting to sudden and unexpected changes in an agent's environment (i.e., novelties). RAPid-Learn is designed to formulate and solve modifications to a task's Markov Decision Process (MDPs) on-the-fly.
    It is capable of exploiting the domain knowledge to learn action executors which can be further used to resolve execution impasses, leading to a successful plan execution.
    We demonstrate its efficacy by introducing a wide variety of novelties in a gridworld environment inspired by {\it Minecraft}, and compare our algorithm with transfer learning baselines from the literature. Our method is (1) effective even in the presence of multiple novelties, (2) more sample efficient than transfer learning RL baselines, and (3) robust to incomplete model information, as opposed to pure symbolic planning approaches.

\end{abstract}

\section{Introduction}

\label{sec:introduction}
AI systems have shown exceptional performance in many ``closed worlds'' domains such as games~\cite{silver2016mastering} where the action space, state space, and the transition dynamics are fixed for duration of the task. Even minor changes to the environment, however, can lead to catastrophic results for closed-world agents~\cite{qu2020minimalistic}. To develop AI systems that can function efficiently in the real world and thus in ``open-world'' settings where sudden and unexpected changes (i.e., novelties) can occur \cite{langley2020open},  we need to relax closed-world assumptions and make agents robust to novel, unseen situations~\cite{kejrivwal2021reinforcing,santucci2020intrinsically}.

The need to adapt to unexpected environmental changes has prompted some to employ {\em Reinforcement Learning} (RL) techniques where learning is lifelong~\cite{abel2018policy,lecarpentier2020lipschitz} and non-stationary~\cite{cheung2020reinforcement}. However, these approaches
assume a continuous evolution of the environment and are less effective
in open-world settings where changes can be abrupt and lead to catastrophic forgetting, thus requiring the agent to potentially re-learn the entire task. 
Symbolic planning approaches, on the other hand, involve an agent reasoning about sequences of predefined operators and optimizing the resultant action sequence that reaches a goal with high
probability~\cite{ghallab_nau_traverso_2016}. These approaches effectively plan
over long horizons, and work well when the domain knowledge
and planning operators are available and defined. However, these approaches are laborious to design and assume an accurate and complete model of the environment. They are often ineffective in evolving, non-stationary, and open-world environments, when the known model is incomplete. 
\begin{figure}[t]
    \centering
    \includegraphics[width=0.35\textwidth]{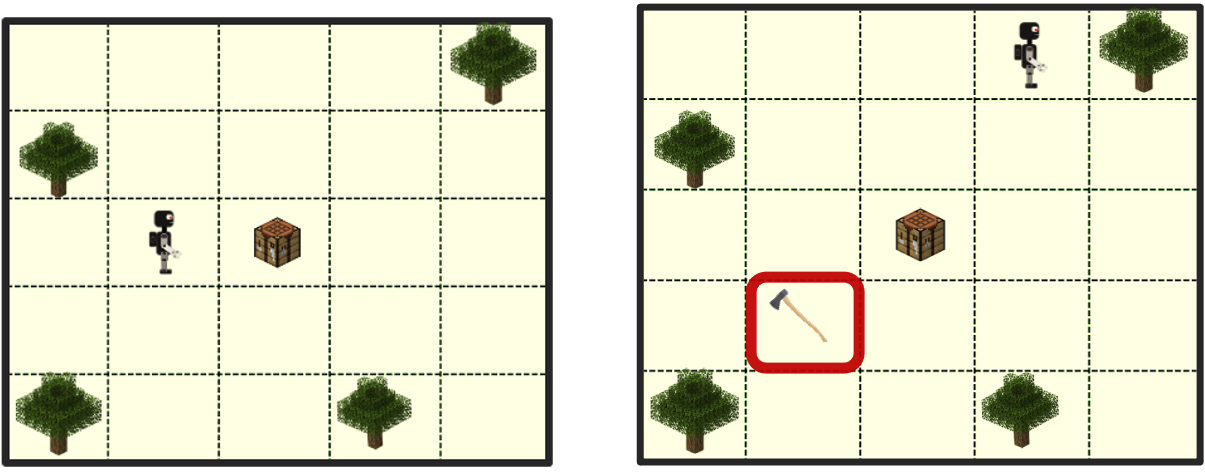}
    \caption{\small Left: Pre-novelty domain showing an agent which can craft items by breaking \emph{trees}. Right: Novelty induces an impasse which requires the agent to use the axe (shown in red) to break the \emph{trees}.}
    \label{fig:running_example_main}
\vspace{-1.5em}
\end{figure}

In this paper we demonstrate that a symbiotic combination of RL-based and planning-based approaches can overcome the shortcomings of either approach and is thus desirable for changing environments (Figure~\ref{fig:running_example_main} shows an example of a changing environment).  However, integrating these paradigms is challenging, because the RL policies need to be abstracted in a way that is beneficial for symbolic planners.  
The challenges are best illustrated by recent approaches aimed at integrating learning and planning which have a limited set of state and action spaces, or are computationally expensive \cite{eppeetal19}.
While employing RL to learn low-level policies for high-level plan operators can help to address the agent's incomplete knowledge in open worlds \cite{Kokel2021,jin2021creativity}, recent hybrid approaches are either not evaluated in open-world settings where plan failures are typical, or
assume a fixed environmental configuration and a custom planner~\cite{sarathy2021spotter}, making them applicable only to a subset of novelties. 

Our proposed solution is RAPid-Learn, a method for recovering from fatal plan failures caused by novelties (sudden and unexpected changes that can result in an execution impasse) (refer Figure~\ref{fig:running_example_main}) by exploiting symbolic knowledge to perform sample-efficient explorations of novelties.  To demonstrate its effectiveness, we compare the novelty adaptation success with state-of-the-art RL methods spanning hierarchical and transfer-learning approaches. A thorough evaluation of our agent in a Minecraft-inspired domain demonstrates the superiority of the proposed methods compared to RL agents.


\section{Related Work}
\label{sec:related_work}
The problem we address has been referred to as ``open-world novelty accommodation''~\cite{langley2020open,muhammad2021novelty}. Here, the aim is to detect and accommodate the novelty, without any prior information of its dynamics.
\begin{figure*}[t]
	\centering
		\centering
		\includegraphics[width=1\textwidth]{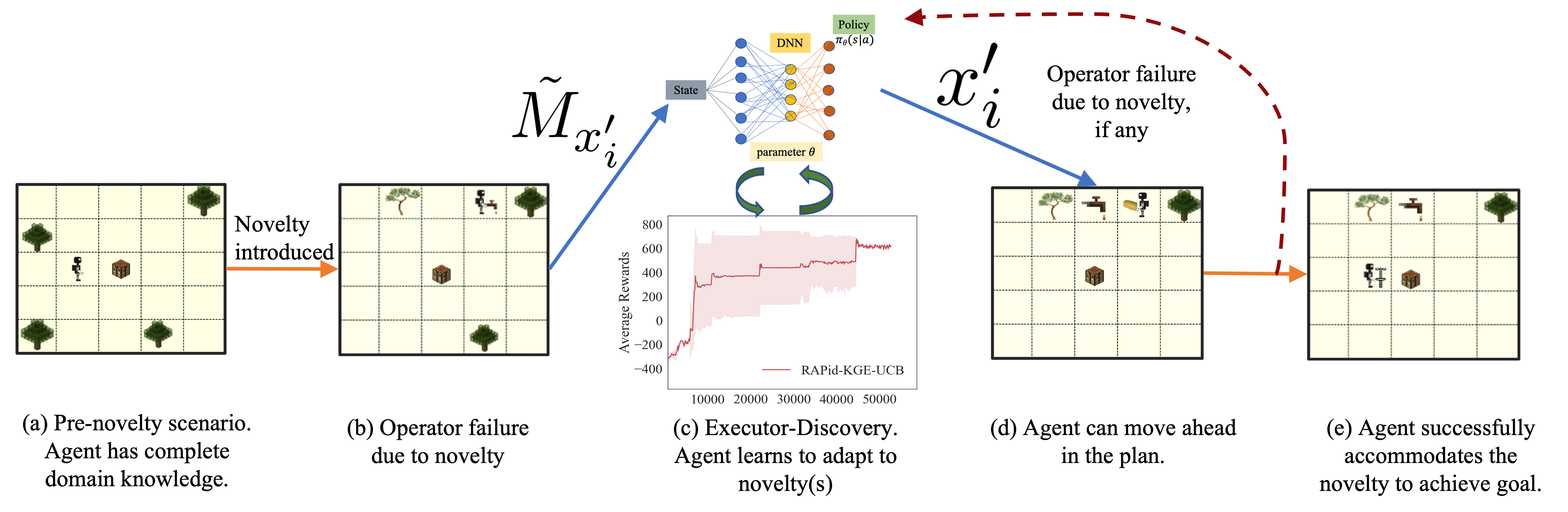}
    \caption{\small Overview of the experimental setup.(a) Illustrates the NovelGridworlds domain in the pre-novelty scenario, showing an agent in a walled arena surrounded by \emph{trees}, and facing a \emph{crafting-table}. (b) Shows the scenario when the environment introduces the \emph{rubber-tree-hard} novelty and the agent reaches an impasse.(c) \emph{Executor-Discovery} phase when the agent accommodates the novelty using our approach RAPid-KGE-UCB. (d) Agent successfully resolves the plan execution impasse. (e) Agent reaches the goal state of crafting a \emph{pogo-stick}.}
	\label{fig:running_example}
\vspace{-1em}
\end{figure*}

\textbf{Symbolic planning} approaches have shown progress in adapting to unforeseen situations in the environment by exploiting existing domain knowledge~\cite{talamadupulaetal17acs,nair2019tool,takahashi2015dynamic}. However, they fail to explicitly accommodate the dynamics of the novelty, while focusing on open-world goals and instructions. Symbolic knowledge representations have also been used for plan recovery~\cite{ouali2015plan}, with one key assumption of an accurate knowledge model. Updating the pre-conditions and effects of the agent's operators is effective in open-world settings~\cite{muhammad2021novelty,gizzi,tungaaai}, although the lack of model-free learning limits novelty handling to only those novelties that can be derived from the existing knowledge base. To bridge this gap, we propose accommodating the novelty by instantiating a model-free reinforcement learner.

\textbf{Reinforcement learning} approaches typically adapt to continuously evolving and non-stationary environments~\cite{khetarpal2020towards,lecarpentier2020lipschitz,padakandla2020reinforcement,cheung2020reinforcement}. Meta-RL attempts at adapting to multiple MDPs, and solving an unseen task drawn from the same distribution as the training tasks~\cite{yu2020meta}. However, these approaches either consider gradually evolving environments or are suited for tasks drawn from the same distribution, and fail to adapt to sudden changes in the environment dynamics~\cite{johnson2022l2explorer}. Transfer in reinforcement learning transfers knowledge from the source task to a complex target task for quick and sample-efficient learning~\cite{JMLR:v10:taylor09a}.
Unlike prior work, we propose a sample-efficient hybrid learner and planner, aimed at detecting and accommodating novelties, when plan execution fails.


\textbf{Hybrid Planning and Learning} approaches take advantage of a sub-symbolic reinforcement learning agent to aid a symbolic planner to address non-stationarity in open-world settings~\cite{sarathy2021spotter,Yang2021}. RL sub-goal learner provides a robust sub-symbolic policy for each operator of a symbolic planner
\cite{Kokel2021, Lyu2019, Illanes2020}. However, they assume an accurate knowledge representation and are not discussed in open-world settings. Other approaches that utilize an RL agent to accommodate the novelty~\cite{sarathy2021spotter} assume a stationary configuration of the initial state, require a custom planner, do not exploit existing domain knowledge for RL exploration, and do not work with function approximators, rendering a sample inefficient hybrid agent catering to only a subset of open-world novelties. To the best of our knowledge, our approach is the first that adapts to the sudden, unknown changes by instantiating a deep reinforcement learner and exploits domain knowledge to speed-up learning.

\section{Preliminaries}
\label{sec:preliminaries}

\subsection{Symbolic Planning}\label{sec:symbolic_planning}

We assume that the agent starts with a domain knowledge, grounded using PDDL~\cite{mcdermott1998pddl}, defined as ~$\Sigma=\langle\mathcal{E}, \mathcal{F}, \mathcal{S}, \mathcal{O}\rangle$, where $\mathcal{E}=\left\{\varepsilon_{1}, \ldots \varepsilon_{|\mathcal{E}|}\right\}$ is a finite set of known entities within the environment. $\mathcal{F}=\left\{f_{1}(\odot), \ldots f_{|\mathcal{F}|}(\odot)\right\}, \odot \subset \mathcal{E}$ is a finite set of known predicates with their negations. Each predicate $f_{i}\left(\odot\right)$, along with its negation $\neg f_{i}(\odot)$, is contained in $\mathcal{F}$. $\mathcal{S}=\left\{s_{1} \ldots s_{|\mathcal{S}|}\right\}$ is the set of symbolic states in the environment. $\mathcal{O}$ denotes the set of known action operators such that $\mathcal{O}=\left\{o_{1}, \ldots o_{|\mathcal{O}|}\right\}$. Each operator $o_i$ is defined with a set of preconditions and effects, denoted $\psi_{i}, \omega_{i} \in \mathcal{F}.$ The preconditions $\psi_{i}$ and effects $\omega_i$ of $o_{i}$ indicate the predicates that must hold true (or false) before and after executing $o_{i}$, respectively. 
We define a planning task as $T=\left(\mathcal{E}, \mathcal{F}, \mathcal{O}, s_{0}, s_{g}\right)$, in which $s_{0} \subset \mathcal{S},$ is the set of starting states and $s_{g} \subset \mathcal{S}$ is the set of goal states. The solution to the planning task $T$ is an ordered list of operators, given by plan $\mathcal{P}=\left[o_{1}, \ldots o_{\mid \mathcal{P}]}\right]$. 


\subsection{Reinforcement Learning \& Sub-Symbolic Executors}
\label{sec:RLandSSExec}

An episodic Markov Decision Process (MDP) $\tilde{M}$ is defined as a tuple $\langle\tilde{\mathcal{S}},\! \tilde{\mathcal{A}},\tilde{\! p},\! \tilde{r},\! \tilde{\gamma} \rangle$, where $\tilde{\mathcal{S}}\!$ is the set of sub-symbolic states, $\tilde{\mathcal{A}}\!$ is the set of actions, $\tilde{p}(\tilde{s}_{t+1}\!|\tilde{s}_t,\tilde{a}_t)\!$ is the transition function, $\tilde{r}(\tilde{s}_{t+1},\tilde{a}_t,\tilde{s}_t)$ is the reward function and $\tilde{\gamma}\!\in\! [0,1]$ is the discount factor. For each timestep $t$, the agent observes a state $\tilde{s}$ and performs an action $\tilde{a}$ given by its policy function $\pi_\theta(\tilde{a}|\tilde{s})$, with parameters $\theta$. The agent's goal is to learn an \emph{optimal policy $\pi^*\!$}, maximizing its discounted return $G_0 = \sum^{K}_{k = 0}\!\tilde{\gamma}^k\! \tilde{r}(\tilde{s}'_k,\tilde{a}_k,\tilde{s}_k) $ until the end of the episode at timestep $K$.

At the sub-symbolic level, we define a set of action executors $\mathcal{X} = \{{x}_1, x_2, x_3, \ldots\}$. Inspired by the options framework in RL~\cite{SUTTON1999181}, an action executor $x_i$ consists of $\langle \mathcal{I}_{x_i}, \pi_{x_i}, \beta_{x_i} \rangle$, where $\mathcal{I}_{x_i} \subseteq \tilde{\mathcal{S}}$ is the initiation set, denoting the set of states when the action executor $x_i$ is available for execution, and it follows a policy $\pi_{x_i}: \tilde{\mathcal{S}} \times \tilde{\mathcal{A}} \rightarrow[0,1]$. $\beta_{x_i}(\tilde{s}_{0},\tilde{s}) \in \{0,1\}$ is the indicator variable 
with value $1$ if $x_i$ can be terminated at $\tilde{s}$ given it was initialised at $\tilde{s}_0$, and $0$ otherwise.

\subsection{Novelty}
\label{sec:novelty}

We define novelty as a completely new encounter for the agent, where the agent can neither derive the dynamics of the novelty using its cognitive abilities, nor through its previous experiences (given domain knowledge $\Sigma$)~\cite{muhammad2021novelty}. Formally, we define novelty as a tuple ~$\mathcal{N}=\langle\mathcal{E'}, \mathcal{F}', \mathcal{S'}, \mathcal{O'}\rangle$, where $\mathcal{E'}$ represents the set of novel entities in the environment such that $\mathcal{E}' \cap \mathcal{E} = \emptyset$. $\mathcal{S'}$ represents the set of novel states such that $\mathcal{S'} \cap \mathcal{S} = \emptyset$, and $\mathcal{O'}$ denotes the set of novel operators such that $\mathcal{O'} \cap \mathcal{O} = \emptyset$, $\mathcal{F}'$ is a set of novel predicates which are unknown to the agent. We assume that $\mathcal{N}$ transforms the domain knowledge from $\Sigma$ to $\Sigma'$ by including novelties which results in an execution impasse. The new domain $\Sigma'$ can generate a planning solution, but the plan execution will result in a failure due to incomplete domain knowledge.



\section{Problem Formulation}
\label{sec:methodology}

\newcommand{\expectedvalue}{\mathbb{E}_{\policy}}
\newcommand{\true}{\mathit{true}}
\newcommand{\false}{\mathit{false}}

\newcommand{\task}{T}
\newcommand{\fluents}{F}
\newcommand{\literals}{L}
\newcommand{\operators}{O}
\newcommand{\initfluentstate}{\sigma_0}
\newcommand{\goal}{\partialfluentstate_g}
\newcommand{\subgoal}{\partialfluentstate_{sg}}
\newcommand{\subgoals}{\Sigma_{sg}}
\newcommand{\fluent}{f}
\newcommand{\literal}{l}
\newcommand{\fluentstate}{\sigma}
\newcommand{\partialfluentstate}{\tilde{\sigma}}
\newcommand{\operator}{o}
\newcommand{\partialoperator}{\tilde{o}}
\newcommand{\plan}{\pi_T}
\newcommand{\precon}{pre}
\newcommand{\effect}{\mathit{eff}}
\newcommand{\addeffect}{\mathit{eff}^+}
\newcommand{\deleffect}{\mathit{eff}^-}
\newcommand{\static}{static}
\newcommand{\unknown}{unknown}
\newcommand{\allliterals}{\mathcal{L}(\fluents)}
\newcommand{\fluenttransition}{\delta(\partialfluentstate, \operator)}
\newcommand{\applyoperator}[2]{\delta(#1, #2)}
\newcommand{\regressoperator}[2]{\delta^{-1}(#1, #2)}
\newcommand{\applyfunction}{\delta}
\newcommand{\regressfunction}{\delta^{-1}}

\newcommand{\reachable}{\Delta_{\partialfluentstate}}
\newcommand{\planstates}{\Sigma_{reach}}
\newcommand{\unplanstates}{\Sigma_{dead}}
\newcommand{\restrict}{restrict}

\newcommand{\planfluentstates}{\Sigma_{plan}}
\newcommand{\reachfluentstates}{\Sigma_{reach}}
\newcommand{\deadfluentstates}{\Sigma_{dead}}

\newcommand{\mdp}{\mathcal{M}}
\newcommand{\mstates}{\mathcal{S}}
\newcommand{\actions}{\mathcal{A}}
\newcommand{\rewardfunction}{R}
\newcommand{\transitionprob}{\tau}
\newcommand{\discountfactor}{\gamma}
\newcommand{\mstate}{s}
\newcommand{\initialstatedist}{\iota}
\newcommand{\initialstate}{\mstate_0}
\newcommand{\action}{a}
\newcommand{\valuefunction}{v_{\policy(\mstate)}}
\newcommand{\policy}{\pi_M}
\newcommand{\policies}{\Pi_M}
\newcommand{\achievable}{\Omega_{\mstate}}
\newcommand{\qfunction}{q(\mstate, \action)}
\newcommand{\qfunctionopt}{q^*(\mstate, \action)}
\newcommand{\qfunctionnext}{q(\newstate, \newaction)}
\newcommand{\learningrate}{\alpha}
\newcommand{\experience}{(\mstate, \action, \rewardfunction, \newstate}
\newcommand{\newstate}{\mstate^\prime}
\newcommand{\newaction}{\action^\prime}
\newcommand{\onlinelearner}{\ell_{expl}}
\newcommand{\offlinelearner}{\ell}
\newcommand{\learners}{L}
\newcommand{\threshold}{\tau}

\newcommand{\exec}{x}
\newcommand{\execinit}{I_x}
\newcommand{\execpolicy}{\pi_x}
\newcommand{\execterm}{\beta_x}
\newcommand{\execs}{X}

\newcommand{\execstar}{x^\star}
\newcommand{\execinitstar}{I_{x^\star}}
\newcommand{\execpolicystar}{\pi_{x^\star}}
\newcommand{\exectermstar}{\beta_{x^\star}}

\newcommand{\execinitfor}[1]{I_{#1}}
\newcommand{\execpolicyfor}[1]{\pi_{#1}}
\newcommand{\exectermfor}[1]{\beta_{#1}}

\newcommand{\explorationpolicy}{\pi_{expl}}

\newcommand{\symdp}{\mathcal{T}}
\newcommand{\detector}{d}
\newcommand{\executor}{e}
\newcommand{\macgyver}{\tilde{\symdp}}

\newcommand{\applicablefluentstates}{\Sigma_{app}}
\newcommand{\beenfluentstates}{\Sigma_{been}}
\newcommand{\abovethresholdfluentstates}{\Sigma_{>\tau}}
\newcommand{\node}{node}
\newcommand{\common}{\partialfluentstate_{common}}
\newcommand{\anotherfluentstate}{\fluentstate^\prime}

\newcommand{\targetmdp}{M_T}
\newcommand{\bsn}{\subgoal}

\newcommand{\lflag}{\mathit{impasse}}

\newcommand{\solve}{\textbf{\textsc{solve}}}
\newcommand{\learn}{\textbf{\textsc{learn}}}
\newcommand{\genprecon}{\textbf{\textsc{gen-precon}}}
\newcommand{\execute}{\textsc{execute}}
\newcommand{\env}{\mbox{ENV}}
\newcommand{\owfs}{\textsc{owfs}}
\newcommand{\owbs}{\textsc{owbs}}

We extend our previous work and develop a framework for integrating planning and learning, in which an integrated planning task allows us to ground operators in an MDP, express goals symbolically, and actualize action hierarchies~\cite{sarathy2021spotter}.
We define an \textit{executor} for a given MDP $\tilde{M}_{x_i}=( \tilde{\mathcal{S}}_{x_i}, \tilde{A}_{x_i}, \tilde{p}_{x_i}, \tilde{r}_{x_i}, \tilde{\gamma}_{x_i} )$ as a triplet $x_i=\langle \mathcal{I}_{x_i}, \pi_{x_i}, \beta_{x_i} \rangle$ where $\mathcal{I}_{x_i} \subseteq \tilde{\mathcal{S}}_{x_i}$ is an initiation set, $\pi_{x_i}(\tilde{a}|\tilde{s})$ is the probability of performing $\tilde{a}$, given current state $\tilde{s}$, and $\beta_{x_i}(\tilde{s}_{0},\tilde{s}) \in \{0,1\}$ is the indicator variable with value $1$ if $x_i$ can be terminated at $\tilde{s}$ given it was initialized at $\tilde{s}_0$, and $0$ otherwise. The policy and termination condition of an executor rely on where it was initiated. 
We define $\mathcal{X}$ as the set of all executors for the set of MDP $\mathcal{M}$.(Section~\ref{sec:RLandSSExec}).
	
	\begin{definition}(Integrated Planning Task)
An \emph{Integrated Planning Task} (IPT) is $\symdp =
	\langle \task, \mathcal{M}, \detector, \executor
	\rangle $ where  $\task = \langle \mathcal{E}, \mathcal{F}, \mathcal{O}, s_o, s_g
	\rangle$ is a STRIPS task, $\mathcal{M}$ is the set of MDPs. A
	detector function $\detector : \tilde{\mathcal{S}} \mapsto \mathcal{S}$ determines a
	symbolic state for a given sub-symbolic MDP state, and an executor function $\executor :
	\mathcal{O} \mapsto \mathcal{X}$ maps an operator to an executor~\cite{sarathy2021spotter}.
\end{definition}
    As described in Sec.~\ref{sec:symbolic_planning}, let $\mathcal{P}$ be the solution of a planning task $T(\mathcal{E},\mathcal{F}, \mathcal{O}, s_0, s_g)$.
	We assume that for each operator $\operator \in \mathcal{O}$, its executor $\executor(\operator)$ accurately maps to $\operator$; that is, for every $\operator \in \operators$, $\mathcal{I}_{\executor(\operator)} \supseteq \{ \tilde{s}_0 \in \tilde{\mathcal{S}}_{e(o)}: \detector(\tilde{s}_0) \supseteq \psi_o \}$ and
	\begin{equation}
	\small
	\exectermfor{\executor(\operator)}(\tilde{s}_0, \tilde{s}) = \begin{cases}
	1 & \textrm{if }(\detector(\tilde{s}) \supseteq \omega_o \vee ((d(\tilde{s}) \supseteq \omega_{{\hat{o}}}) \forall \hat{o} \in \hat{\mathcal{O}}) \\ & \quad \wedge ~\exists\mathcal{\hat{P}}\\
	0 & \textrm{otherwise}.
	\end{cases}
	\label{eq:indicator_function}
	\end{equation}
The executor $e(o)$ reaches a termination state $\tilde{s}$ when it satisfies the effects of the operator $o$ ($d(\tilde{s}) \supseteq \omega_o$), or if it satisfies the effects of all subsequent operators $\hat{o}\in\hat{\mathcal{O}}$ in the plan $\mathcal{P}$ (($d(\tilde{s}) \supseteq \omega_{\hat{o}}) \forall \hat{o} \in \hat{\mathcal{O}} $), where the preconditions of $\hat{o}$ contain the effects of $o$ ($\psi_{\hat{o}} \supseteq \omega_o$). Also, ensuring a planning solution $\hat{\mathcal{P}}$ to the task $\hat{T} = (\mathcal{E}, \mathcal{F}, \mathcal{O}, d(\tilde{s}), s_g)$ exists. A \emph{solution} to the IPT $\mathcal{T}$ is an ordered list of executors $\left[x_{1}, \ldots x_{\mid \mathcal{X}]}\right] \in \mathcal{X}$ having the above mentioned properties. A \emph{planning solution} to an IPT $\mathcal{T}$ is the ordered list of operators given by the $\mathcal{P}=\left[o_{1}, \ldots o_{\mid \mathcal{P}]}\right]$. Executing the ordered list of executors in the sub-symbolic space will yield a final state $\tilde{s}$ such that $d(\tilde{s}) \supseteq s_g$, thus reaching the goal state $s_g$.
$\symdp$ is solvable if a solution exists and plannable if a plan exists.
		
\subsection{The Executor Discovery Problem}
We define a stretch-IPT to capture hard but feasible goals that require the discovery of missing executors.
\begin{definition}
	(Stretch-IPT). A Stretch-IPT $\macgyver$ is an
	IPT $\symdp$ for which a solution exists, but a planning solution does not.
\end{definition}
A novelty introduced by the environment, results in incomplete domain knowledge (Section~\ref{sec:novelty}), causing an execution impasse due to operator failure. We are interested in finding a solution to the stretch-IPT, specifically study how to automatically generate missing executors on-the-fly to solve the execution impasse. 

\begin{definition}
(Executor Discovery Problem). Given a stretch-IPT $\macgyver = \langle \task', \mathcal{M}', \detector',
\executor' \rangle $ with $\task'=\langle \mathcal{E} \cup \mathcal{E}', \mathcal{F}\cup \mathcal{F}', \mathcal{O} \cup \mathcal{O}', s'_o, s_g
\rangle$, construct a set of executors $\{x'_{1}, \ldots, x'_{m}\} \in \mathcal{X}$ for the set of failed operators $\{o_{1}, \ldots, o_{m}\} \in \mathcal{O} $ such that the stretch-IPT $\macgyver$ is solvable, with  the executor function: 
\begin{equation} \executor'(o_{i}) =  x'_{i}, \operator_i \notin \mathcal{O}
\end{equation}
i.e., we find an executor whose operator does not exist in $\mathcal{O}.$
\end{definition}

\section{RAPid-Learn}
Below we describe a running example followed by a detailed description of our approach.
\subsection{Description of the domain}
\label{sec:domain_desc}
Figure~\ref{fig:running_example}a shows a gridworld representation of a \emph{Minecraft} inspired domain. An agent is depicted confined in a walled arena that contains \emph{trees} and a \emph{crafting-table}. The agent can obtain \emph{tree-log} by breaking {\em trees}. \emph{Tree-logs} can be crafted into \emph{planks}, which in turn can be crafted into \emph{sticks}. A combination of \emph{planks} and \emph{sticks} can be crafted into a \emph{tree-tap} when facing the \emph{crafting-table}.~\emph{Rubber} can be extracted, by selecting \emph{tree-tap} in front of a \emph{tree}. 
Finally, a \emph{pogostick} can be crafted using a combination of \emph{sticks}, \emph{planks} and \emph{rubber}.

\subsection{Running Example}
\label{sec:running_example}
Let us consider a novelty scenario in which a \emph{rubber-tree} ($\varepsilon_1 \in \mathcal{E}'$) appears in the environment (Figure~\ref{fig:running_example}b). The novel environment also provides a new action \emph{place-tree-tap}, and a novel action executor  \emph{approach-rubber-tree}. The agent can only extract \emph{rubber} by placing a \emph{tree-tap} in front of the \emph{rubber-tree}. This results in the failure of operator \emph{extract-rubber} $(o_i)$. To overcome this execution impasse, an MDP $\tilde{M}_{x_i}$~(Figure~\ref{fig:running_example}(c)) is instantiated, whose solution is the learned executor $x'_i$ which is mapped to the failed operator $o_i$ for future use. Action executor $x'_i$'s policy $\pi_{x'_i}$ will consists of steps that involve the agent to stand one block away from the \emph{rubber-tree}, place the \emph{tree-tap} next to the \emph{rubber-tree} and then use the failed operator \emph{extract-rubber}. After the agent successfully executes $x'_i$ (Figure~\ref{fig:running_example} (d)), it switches back to original plan to craft a \emph{pogostick} (Figure~\ref{fig:running_example}(e)).



\subsection{Running in an impasse}

As shown in Algorithm~\ref{alg:rapidlearn_algorithm}, the agent starts with a stretch integrated planning task $\mathcal{\tilde{T}}$ as an input. Using the domain knowledge grounded through PDDL~\cite{mcdermott1998pddl}, the agent uses MetricFF~\cite{hoffmann2003metric} planner to generate a planning solution (plan $\mathcal{P}$) to the Stretch-IPT $\mathcal{\tilde{T}}$ (line 3).
The agent then executes this plan (line 4-8) in a novel environment. We assume that one of the operators will fail in the novel environment, resulting in an execution impasse. The impasse occurs if the known effects of the operator are not true after the execution of the operator in the environment.
Once the agent detects an execution impasse, it checks if it has a corresponding sub-symbolic action executor $x'_i$ for the failed operator $o_i$ (line 10-12). If the executor does not exist, the agent enters \textit{executor-discovery} mode (line 13). We now describe how the agent instantiates and solves an executor discovery problem to succeed in the execution impasse induced by the novelty.

%
\begin{algorithm}[tb]
\footnotesize
\caption{~\textit{\textbf{RAPid-Learn(~$\mathcal{\tilde{T}}$)}}}
\label{alg:rapidlearn_algorithm}
\begin{algorithmic}[1] 
\STATE $\macgyver = \langle \task', \mathcal{M}', \detector', \executor' \rangle$: Stretch Integrated Planning Task
\STATE $\task'=\langle \mathcal{E} \cup \mathcal{E}', \mathcal{F}\cup \mathcal{F}', \mathcal{O} \cup \mathcal{O}', s'_o, s_g
\rangle$
\STATE {$\mathcal{P}$ = \textbf{Planner}($\mathcal{\tilde{T}}$)}~~~~\COMMENT{$\mathcal{P}=\{ o_1, o_2, ..., o_{|\mathcal{P}|}\}$}
\FOR{$o_i \in \mathcal{P}$}
\STATE $\mathcal{X} \leftarrow \mathcal{X} \cup e(o_i)$
\ENDFOR
\FOR {$o_i \in \mathcal{P}$}
\STATE Success = \textbf{Execute} ($o_i$)
\IF {$\neg$Success}
\IF{$x'_i \in \mathcal{X}$}
\STATE {\textbf{Execute}($x'_i$)}
\ELSE
\STATE ${x'_i} \leftarrow$  \textit{\textbf{Discover-Executor($o_i, \mathcal{P}, \tilde{\mathcal{T}}, k$)}} ; $\mathcal{X} \leftarrow \mathcal{X}\! \cup\! \{x'_i\}\!$
\STATE $\textbf{Execute}(x'_i)$
\ENDIF
\ENDIF
\ENDFOR
\end{algorithmic}
\end{algorithm}
%

%
\begin{algorithm}[!tbh]
\footnotesize
\caption{~\textit{\textbf{Discover-Executor($o_i$, $\mathcal{P}$, $\tilde{\mathcal{T}}$, $k$) $\rightarrow$ $x'_i$}}}
\label{alg:learn_algorithm}
\begin{algorithmic}[1] 
\STATE $\macgyver = \langle \task', \mathcal{M}', \detector',
\executor' \rangle $; $\task'=\langle \mathcal{E} \cup \mathcal{E}', \mathcal{F}\cup \mathcal{F}', \mathcal{O} \cup \mathcal{O}', s'_o, s_g
\rangle$
\STATE Hyperparameters : $\rho_0, \epsilon_0, c, \mu, e_{max}$
\STATE $\tilde{\mathcal{S}'_{x'_i}}:$ set of sub-symbolic states;  $\tilde{\mathcal{A}}'_{x'_i}:$ set of actions; 
\STATE $\beta_{x'_i}:$ {indicator function for executor $x'_i$ as defined in \ref{eq:indicator_function}}
\STATE  $\tilde{{\mathcal{S}_0}} \in \tilde{\mathcal{S}'_{x'_i}}$  set of initial states; $\Delta \gets \emptyset:$~set of actions to bias
\STATE $\mathcal{E}_{novel} \gets \{\mathcal{E}'\} \setminus \{\mathcal{E}\}$: set of novel entities
\STATE $\tilde{\mathcal{A}}_{novel}\gets \{\tilde{\mathcal{A}}'_{x'_i}\} \setminus \{\tilde{\mathcal{A}}\}$: set of novel actions
\STATE $\Delta \leftarrow \Delta \cup \{\tilde{\mathcal{A}}_{novel}, o_i\}$
\STATE $\text{StateList}  \leftarrow [\:], \text{ActList}  \leftarrow [\:], \text{DoneList}  \leftarrow [\:], \text{RewList} \leftarrow [\:]$
\STATE $\mathcal{S}_r \gets $ \textit{\textbf{PlannableStateGenerator}$(\Sigma, \mathcal{P}, o_i$)}
\STATE $e \leftarrow 1:$ episode number
\WHILE {$e<e_{max}$}
\STATE $\tilde{s}'_0 \leftarrow:$ \textit{\textbf{ReachFailedOperator}}$(\tilde{\mathcal{T}})$: initial state
\STATE $done \leftarrow$ \textit{false}, $t \leftarrow 1$
\IF {$k \neq$ {\textit{EG}} $\!\And\! uniformRandomNo(0,1) < \rho_0$}
\STATE $\varepsilon \leftarrow$ \textit{random}$(\mathcal{E}_{novel})$ ; $p \leftarrow$ \textit{Planner}$(\varepsilon, \tilde{s}'_0)$
\STATE $\tilde{s}'_t \leftarrow$ {\textit{execute}}$(\tilde{s}'_0, p)$
\ENDIF
\STATE $A_{p} \leftarrow$ $\{ \pi(\tilde{a}'_i|\tilde{s}'_t)\:\forall\: \tilde{a}'_i \in \tilde{\mathcal{A}'_{x'_i}} \}$
\WHILE {$\neg $\textit{done} ~$\lor~ t<U$}
\IF{$uniformRandomNo(0,1) < \epsilon_0$}
\IF {$k$ = \textit{KGE-UCB}}
\FOR{$\tilde{a}' \in \tilde{\mathcal{A}}'_{x'_i}$}
\STATE \small{$A_{p}[\tilde{a}'] \leftarrow A_{{p}}[\tilde{a}']\left\{\begin{array}{l}\!\!+c \sqrt{\frac{\ln t}{N_{t}(\tilde{a}')}} \hspace{0.35cm}\!\!\!\textit{if }\! \tilde{a}'\! \in \!\Delta \vspace{0.1cm} \\\!\! -c \sqrt{\frac{\ln t}{N_{t}(\tilde{a}') }} \hspace{0.35cm} \!\!\!\textit{if }\! \tilde{a}' \!\notin \!\Delta\end{array}\right.$}
\ENDFOR
\ELSIF{$k$ = \textit{KGE-UAB}} \vspace{0.1cm}
\FOR{$\tilde{a}' \in \tilde{\mathcal{A}}'_{x'_i}$}
\STATE \small{$A_{p}[\tilde{a}'] \leftarrow \left\{\begin{array}{cc}\frac{(\mu-1+\Sigma \Delta) \tilde{a}'}{\mu \Sigma \Delta} & \textit{if }  \tilde{a}' \in \Delta \vspace{0.1cm}\\ \tilde{a}'/ \mu & \textit{if }  \tilde{a}' \notin \Delta\end{array}\right.$}
\ENDFOR
\ENDIF
\ENDIF
\STATE $\tilde{a}'_t \leftarrow \text{arg-max}(A_p)$
\STATE $\tilde{s}'_{t+1} \leftarrow $ \textit{execute}$(\tilde{a}'_t)$
\STATE $done, r \leftarrow$ \textit{r}$(\mathcal{S}_r,\tilde{s}'_t, \tilde{a}_t, \tilde{s}'_{t+1})$

\STATE $\text{DoneList}.append(done),\: \text{StateList}.append(\tilde{s}'_t)$
\STATE $\text{RewList}.append(r),\: \text{ActList}.append(A_p)$

\STATE $t\leftarrow t+1$, $\tilde{s}'_{t+1}\leftarrow\tilde{s}'$
\ENDWHILE
\STATE $e\leftarrow e+1$
\STATE $\pi \leftarrow$ \textit{Update-Network}$ (\text{StateList}, \text{ActList}, \text{RewList}$)
\IF{ $\textit{Converge}(\text{DoneList}, \text{RewList}) > \eta$}
\STATE $\pi^{c}_{x'_i} \!\leftarrow \!\pi$, $x'_i\! \leftarrow\! \langle \tilde{S}'_0,\pi^{c}_{x'_i},\beta_{x'_i} \rangle$
\RETURN $x'_i$
\ENDIF
\ENDWHILE
\STATE $x'_i \leftarrow \langle \tilde{S}'_0,\pi_{x'_i}, \beta_{x'_i} \rangle$
\RETURN $x'_i$
\end{algorithmic}
\end{algorithm}

\subsection{Executor-Discovery}
To find a solution to the stretch-IPT $\mathcal{\tilde{T}}$, the agent needs to discover an executor $x'_i$ that will succeed through the impasse caused by the novelty. The executor discovery process is described in Algorithm~\ref{alg:learn_algorithm}. The agent instantiates an online reinforcement learner over the episodic MDP $\tilde{M}_{x'_i}=( \tilde{\mathcal{S}}'_{x'_i}, \tilde{A}'_{x'_i}, \tilde{p}'_{x'_i}, \tilde{r}_{x'_i}, \tilde{\gamma}_{x'_i} )$ the first time it encounters the failed operator. This MDP consists of the set of sub-symbolic states $\tilde{\mathcal{S}'_{x'_i}}$ and actions $\tilde{\mathcal{A}'_{x'_i}}$ (sub-symbolic pre-novelty actions, sub-symbolic novel actions, and all the symbolic operators mapped to action executors). A sparse reward function is generated on-the-fly to guide the agent to discover the set of plannable states $\mathcal{S}_r\!$ from where it can reach the goal state $\!s_g$.

\subsubsection{Discovery of Plannable States}
\label{sec:plannable_state_disc}
With the pre-novelty domain knowledge, the agent accumulates the set of preconditions $\Psi$ and effects $\Omega$ of all known operators. The agent then generates a set of plannable states $\mathcal{S}_r$, which contains: 1) the states that satisfy the effects  $\omega_{o_i}$ of the failed operator $o_i$ and 2) the states that satisfy the effects of all subsequent operators $\hat{o}\in\hat{\mathcal{O}}$ in the plan $\mathcal{P}$ (($d(\tilde{s}) \supseteq \omega_{\hat{o}}) \forall \hat{o} \in \hat{\mathcal{O}} $), where the preconditions of $\hat{o}$ contain the effects of $o$ ($\psi_{\hat{o}} \supseteq \omega_o$). In principle, the agent performs an OR operation over 1 and 2 to find a plannable state.\footnote{Further details of the \emph{PlannableStateGenerator} algorithm in Appendix Section ~\ref{sec:additional_algo} }
In each episode, the agent computes a plan from the initial environment configuration, and carries out this plan until it reaches the failed operator (Algorithm 2 - line 13).
The agent is rewarded with a positive reward $\phi_1$, if it reaches a state in $\mathcal{S}_r$ within a predetermined number of timesteps $U$, and a plan from this state to the goal state $s_g$ exists. However, in some cases the agent can reach a state in $\mathcal{S}_r$ but the planning solution from this state to the goal state $s_g$ does not exist (because of irreversible actions taken by the agent). To prevent this failure, a negative reward $\phi_2$ is given, and the episode terminates. The agent gets a unit negative reward for all other steps. We define the reward function as:
%
$$
\small{
{\text{r}}\left(\mathcal{S}_r, \tilde{s}'_t, \tilde{a}'_{t}, \tilde{s}'_{t+1}\right)=\left\{\begin{array}{cl}
\phi_1 & {if}~~ d(\tilde{s}'_{t+1}) \subset \mathcal{S}_r \wedge \exists\hat{{\mathcal{P}}}\\
-\phi_2 & {if}~~ d(\tilde{s}'_{t+1}) \subset \mathcal{S}_r \wedge \nexists {\hat{\mathcal{P}}} \\
-1 & {\text{otherwise}} \\
\end{array}\right.}
$$
\vspace{-1em}
\subsubsection{Knowledge-guided-exploration}
To guide the agent to explore efficiently and utilize domain knowledge, we employ knowledge-guided exploration. We describe two approaches:

{\em Knowledge-guided curriculum learning}: Learning to solve complex problems may require extensive interactions with the environment. Knowledge obtained through simpler sub-tasks can be utilized to reduce the exploration of complex tasks~\cite{narvekar2020curriculum}. \em knowledge-guided curriculum learning \em enables the agent to solve simpler sub-tasks, thereby increasing the probability of the agent landing into a plannable state. With a probability $\rho$, the agent is provided a curriculum to reach a novel state (line 14-16, Alg.~\ref{alg:learn_algorithm}). Randomly selecting a novel entity from the set of novel entities, the agent utilizes the planner to reach this novel entity~(line 15). This procedure enables the agent to begin exploring from a promising initial state ~\cite{narvekar2016source}. The agent is given the state transitions $(\tilde{s}'_t,\tilde{a}'_t,\tilde{s}'_{t+1})$ from its initial state to this promising state.

 {\em Knowledge-guided action biasing}: here, during exploration the agent biases a subset of actions through domain knowledge. The subset of the actions to bias $\Delta$ consists of; (1) novel actions, and (2) failed operators in the form of executors (lines 6-7, Algorithm~\ref{alg:learn_algorithm}). With a probability $\epsilon_0$, the probability of selecting these actions is bumped up when the agent explores the environment (lines 20-26). We develop two methods of action probability bumping, Upper Confidence Bounds (KGE-UCB), and Uniform Action Biasing (KGE-UAB).
In KGE-UCB, we bump up the probability of actions in the set $\Delta$  ($A_p$, line 18) inversely proportional to square root of the number of times the action was executed prior to the current timestep~\cite{auer2002using} (lines 21-24). This helps in selecting those actions that were tried the least, enabling the agent to visit new states and increasing the possibility of reaching the plannable states in $\mathcal{S}_r$. 
In KGE-UAB, we uniformly bump up the selection probability of all the actions in the set $\Delta$ (lines 25-29). Both these methods are an extension of the $\epsilon$-greedy exploration (EG) strategy. Contrary to EG, they exploit domain knowledge, when deciding which action to choose during exploration to perform efficient exploration. 


 The rationale behind using knowledge-guided-exploration in such a way follows two assumptions: 1) The agent should use domain knowledge to guide exploration; 2) The novelties are the reason why the agent gets stuck in an impasse. 
 

The agent continues updating its policy until it reaches the maximum permitted episodes ($e_{max}$) (lines 31-39) or if it converges to a policy $\pi^c_{x'_i}$ defined by a pre-determined success rate convergence threshold $\eta$ (lines 40-42). The agent then exits the executor discovery mode and continues with its original plan $\mathcal{P}$.

\subsection{Recovery}
\label{sec:recovery}
Once the agent learns a policy $\pi_{x'_i}^c$ to succeed at the impasse, it is added as an action executor $x'_i$ (Alg~\ref{alg:learn_algorithm} lines 41,45) to the set of action executors $\mathcal{X}$ (line 10, Alg~\ref{alg:rapidlearn_algorithm}). The agent executes $x'_i$, whenever the operator $o_i$ fails. This framework is effective even in multiple novelty settings, where more than one operators fail. 


\section{Evaluation and Results}
\label{sec:evaluation_and_results}
\subsection{NovelGridworlds Domain}
\label{sec:default_env}

We evaluate RAPid-Learn on a $[12 \times\! 12]$ NovelGridworlds domain~\cite{goel2021novelgridworlds}, a gridworld crafting problem inspired by Minecraft (Target task described in Section~\ref{sec:domain_desc}). A local view of the environment represents the sub-symbolic state of the RL learner, implemented as a LiDAR-like sensor that emits beams for every entity in the environment at incremental angles of \( \frac{\pi}{4} \) to determine the closest entity in the angle of the beam, in other words, LiDAR-like sensor provides observation of size $8\times|\mathcal{E}|$, where $\mathcal{E}$ is the entire set of possible entities. Additional sensors observe the content of the agent's inventory and the currently selected item. The action space is the sub-symbolic action space given by the environment (navigation actions- \emph{turn-left}, \emph{turn-right}, \emph{move-forward}; interaction actions - \emph{break}, \emph{extract-rubber}; crafting-actions - \emph{craft-planks}, \emph{craft-stick}, \emph{craft-pogostick}), augmented with novel actions (according to the novelty, shown in Table~\ref{tab:novelties}) and hierarchical action operators (\emph{approach-entity}, parameterized by \emph{entity}) implemented by the planner. The positive reward constant $\phi_1 = 1000$, and the negative reward constant $\phi_2 = -350$. The RL implemented in the RAPid-Learn architecture uses the policy gradient~\cite{williams1992simple} algorithm. The domain was chosen for two key reasons: 1) it provides a complicated task that involves a sequential set of actions to reach the goal state. If the agent misuses its resources, it will not succeed in the task; and 2) the domain is designed specifically for open-world problem solving, enabling us to create and experiment with a variety of novelties~\cite{goel2021novelgridworlds}.

\begin{table}[t]
\footnotesize
    \begin{center}
    \begin{tabular}{p{1.3cm}p{0.8cm}p{1.3cm}p{3.7cm}}   \arrayrulecolor{taupegray}\midrule
     

\centering \textbf{Novelty} &\centering  \textbf{Entity ($\mathcal{E'}$)} & \centering  \textbf{Operator ($\mathcal{O'}$)} & \centering \textbf{Dynamics} \tabularnewline \midrule

      \centering \textit{axe-to-break (ATB)} [easy/hard] & \centering \textit{axe}  & \textit{select-axe}, \textit{approach-axe[hard]}   & break \textit{tree} only if holding \textit{axe}; \textbf{\em [easy] \em}\textit{axe} present in \textit{inventory}; \textbf{\em [hard] \em }\textit{axe} present in environment\tabularnewline \midrule
     
     
     \centering \textit{fire-crafting-table (FCT)} [easy/hard] &\centering  \textit{fire}, \textit{water} & \textit{select-water}; \textit{spray}, \textit{approach-water[hard]} &  \textit{crafting-table} set on fire; need to \textit{spray} \textit{water} to access \textit{crafting-table}; \textbf{\textit{[easy]}} \textit{water} present in \textit{inventory} \textbf{\em [hard] \em} \textit{water} present in the environment\\ \midrule
     
     
     \centering \textit{rubber-tree (RT)} [easy/hard] &\centering  \textit{rubber-tree} & \centering \textit{place-tree-tap, approach-rubber-tree[hard]} & \textbf{\textit{[easy]}} \textit{rubber} can only be extracted when facing \textit{rubber-tree}; \textbf{\textit{[hard]}} need to place \textit{tree-tap} in front of a \textit{rubber-tree} to extract \textit{rubber}. \\ \midrule
     
     \centering \textit{scrape-plank} & \centering - & \centering \textit{scrape-plank} & cannot obtain \textit{tree-log} by breaking \textit{tree}, can only scrape planks from \textit{tree}\\ 
     \bottomrule
    \end{tabular}
    \caption{\small Novelty descriptions: A novelty changes the environment by adding new entities, operators, and dynamics.}
    \label{tab:novelties}
    \end{center}
\end{table}
\subsection{Experimental Setup}
Our setup (Figure~\ref{fig:running_example}) is designed to evaluate: (1) \textbf{Novelty accommodation}: The agent's ability to solve the impasse, regardless of the nature of novelty. For this, we evaluate RAPid-Learn on a variety of novel scenarios (Table~\ref{tab:novelties}). The scenarios comprise of adding a novel entity, and/or, a novel operator in the agent's environment. In some cases, the agent has to just learn to use the novel operator and the entity, while in others, it has to explore the environment through sub-symbolic actions to solve the impasse. This comprehensive list, though non-exhaustive, captures many aspects of different novelty attributes\footnote{Further details of the implemented novelties are described in Appendix Section~\ref{app:novelty_desc}}(2) \textbf{Sample Efficiency}: The number of interactions taken by the agent to solve the impasse. To compare sample efficiency, we evaluate RAPid-Learn's three variations, namely, 1) RAPid-KGE-UCB: RAPid-Learn with knowledge-guided-exploration using upper confidence bounds; 2) RAPid-KGE-UAB: RAPid-Learn with knowledge-guided-exploration using uniform action biasing; 3) RAPid-EG: RAPid-Learn with $\epsilon$-greedy exploration function, and compare these with two baselines. 


\subsubsection{Baselines}
\label{sec:baselines}
Existing methods in integrating planning and learning are aimed at either generating policies for the high-level operators~\cite{Kokel2021} or formulating operators to tackle the change in the environment~\cite{sarathy2021spotter}. Approaches that generate policies for high-level operators are not discussed in open-world novelty settings, and others that formulate operators~\cite{sarathy2021spotter} are not robust to varying environment configurations and numeric predicates. We therefore compare our architecture against policy reuse~\cite{glatt2017policy} and actor-critic transfer learning~\cite{clegg2017learning}.
For each transfer learning baseline, we pre-train the agent (until convergence:$>\!96\%$ success rate for last 100 episodes) using dense reward shaping, where choosing the right action at the right time is given intermediate reward. 
During pre-novelty training, the observation and the action spaces are extended with placeholder elements to accommodate the introduced novelties that can extend the shape of these spaces. On the pre-trained expanded model, we perform transfer learning using two approaches: 
Policy reuse~\cite{glatt2017policy} transfers the learned policy and Actor-critic transfer~\cite{clegg2017learning} transfers the policy and value function through PPO~\cite{schulman2017proximal}. 

\begin{figure*}[h]
	\centering
	\begin{minipage}{0.49\columnwidth}
		\centering
		\includegraphics[width=\textwidth]{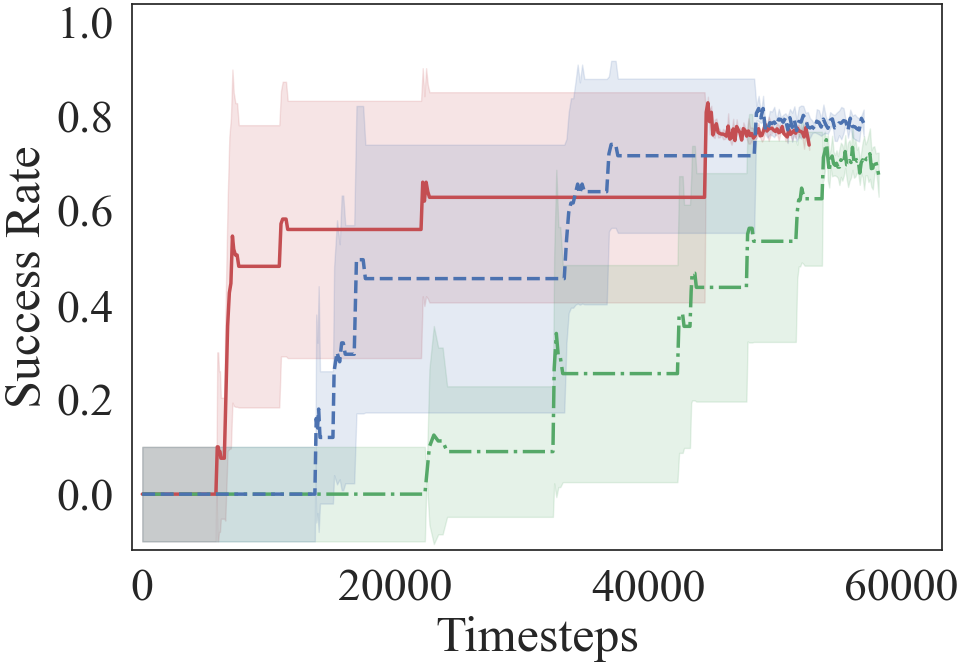}
        \small (a) Rubber Tree Hard
		\label{fig:rthard}
	\end{minipage}%
	\begin{minipage}{0.49\columnwidth}
		\centering
		\includegraphics[width=\textwidth]{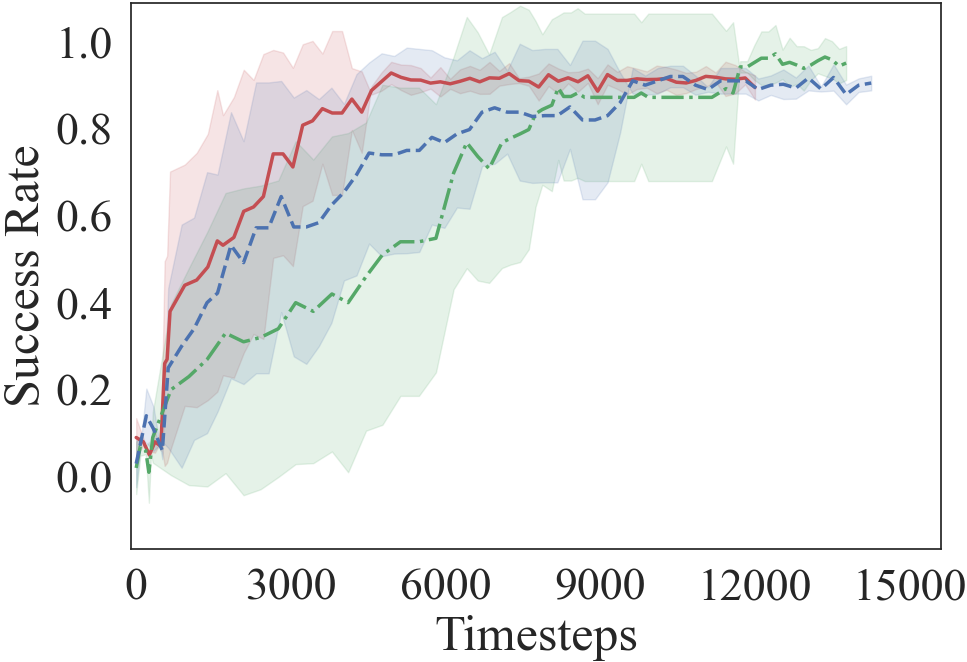}
        {\small (b) ATB+FCT Easy}
		\label{fig:atbfcteasy}
    	\end{minipage}
	\begin{minipage}{0.49\columnwidth}
		\centering
		\includegraphics[width=\textwidth]{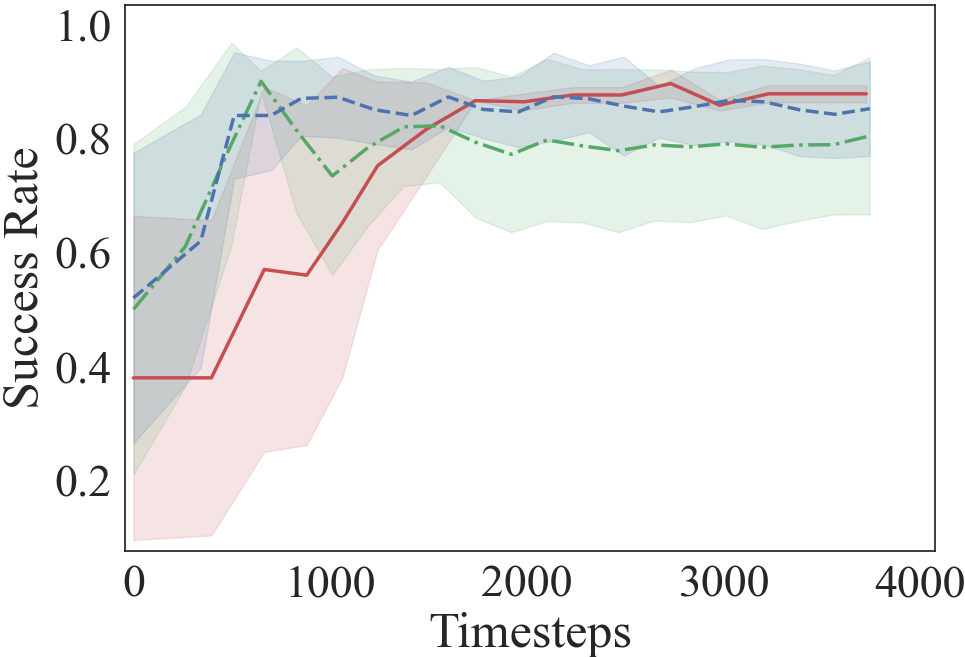}
        \small {(c) Axe to Break Hard}
		\label{fig:atbhard}
	\end{minipage}
	\begin{minipage}{0.49\columnwidth}
		\centering
		\includegraphics[width=\textwidth]{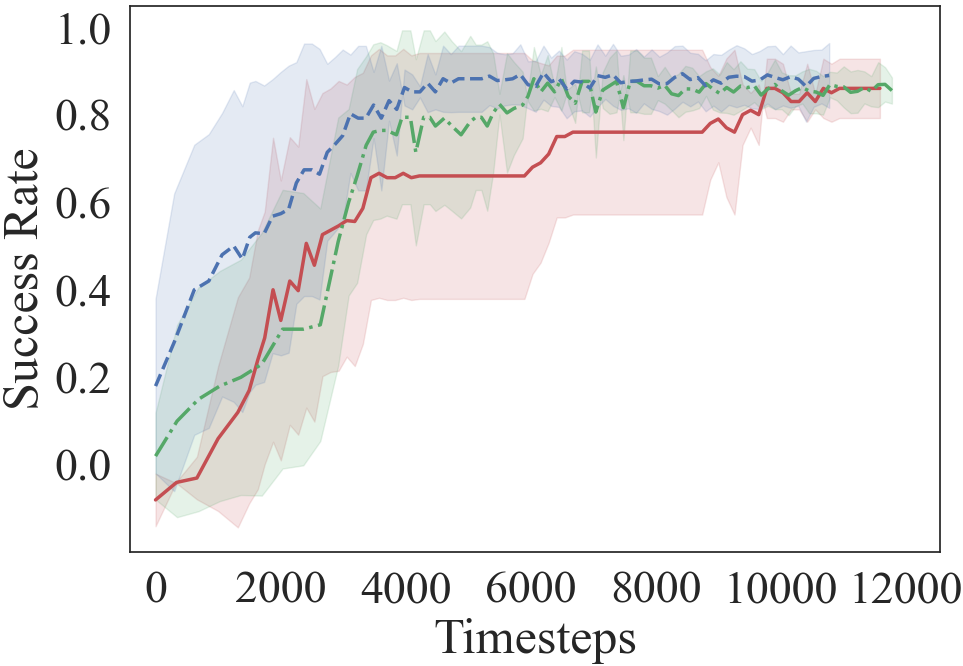}
        \small{(d) Fire Crafting Table Hard}
		\label{fig:fcthard}
	\end{minipage}
	\begin{minipage}{.95\columnwidth}
	\vspace{1.25em}
		\includegraphics[width=1\textwidth, height=0.085\textwidth]{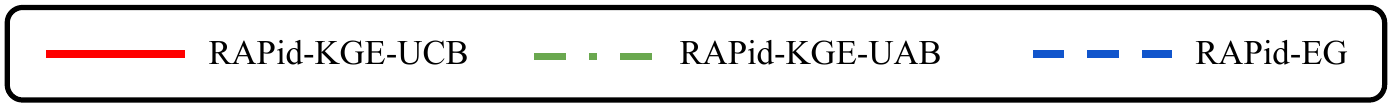}
		\label{fig:graph_label}
	\end{minipage}	
	\vspace{-1.25em}
	\caption{\small 
	These plots illustrate the performance of the proposed RAPid-Learn methods during the \emph{Executor-Discovery} phase of the associated novelty. The baseline methods take orders of magnitude more timesteps to converge, and are shown in Appendix Figure~\ref{fig:baselines_plots}}
\label{fig:plots}
\vspace{-1em}
\end{figure*}

\begin{table}[!tb]
\footnotesize
\begin{center}
\begin{tabular}{>{\centering}p{2.4cm}>{\centering}p{2.6cm}>{\centering}p{2.0cm}p{0.1cm}}
\arrayrulecolor{taupegray}\toprule

\textbf{\begin{small}Agent\end{small}} & {\textbf{\begin{small}Time to adapt (timesteps)\end{small} }} & \textbf{\begin{small}Post-novelty performance (success rate)\end{small} }\tabularnewline
\midrule
 &\begin{footnotesize} Mean$\pm$SD\end{footnotesize}
 & \begin{footnotesize}Mean$\pm$SD\end{footnotesize}
\tabularnewline
\midrule
 \multicolumn{3}{c}{\textbf{Rubber Tree (RT) - Hard}} \tabularnewline
 \midrule
 
\rowcolor{gainsboro}
RAPid-KGE-UCB  &  $(2.85 \pm 2.15)\times10^4$ &\centering $0.95 \pm 0.01$\tabularnewline
RAPid-KGE-UAB  & $(3.28 \pm 1.51) \times10^4$ &\centering $0.93 \pm 0.03$\tabularnewline
RAPid-EG  & $(3.73 \pm 2.08)\times10^4$ &\centering $0.91 \pm 0.04$\tabularnewline
Actor-Critic Transfer  & did not converge &\centering N.A \tabularnewline
Policy-Reuse  & did not converge &\centering N.A \tabularnewline
 \midrule
 \multicolumn{3}{c}{\textbf{Axe to Break (ATB) + Fire Crafting Table (FCT) - Easy}} \tabularnewline
 \midrule
 
\rowcolor{gainsboro}
RAPid-KGE-UCB  & $(4.02 \pm 1.82)\times10^3$ &\centering $0.97 \pm 0.01$\tabularnewline
RAPid-KGE-UAB  & $(4.56 \pm 3.43)\times10^3$ &\centering $0.97 \pm 0.02$\tabularnewline
RAPid-EG  & $(6.93 \pm 4.41)\times10^3$ &\centering $0.98 \pm 0.01$\tabularnewline
Actor-Critic Transfer  & $(17.6\pm 5.13) \times10^4$ &\centering $0.93 \pm 0.02$ \tabularnewline
Policy-Reuse  & $(8.08\pm 2.38)\times10^5$ &\centering $0.92 \pm 0.04$ \tabularnewline

 \midrule
 \multicolumn{3}{c}{\textbf{Axe To Break (ATB) - Hard}} \tabularnewline
 \midrule
 
\rowcolor{gainsboro}
RAPid-KGE-UCB  & $(2.35 \pm 1.27)\times10^3$ &\centering $0.95 \pm 0.01$\tabularnewline
RAPid-KGE-UAB  & $(13.0 \pm 4.89)\times10^2$ &\centering $0.85 \pm 0.07$\tabularnewline
RAPid-EG  & $(14.9\pm 8.65)\times10^2$ &\centering $0.84 \pm 0.03$\tabularnewline
Actor-Critic Transfer  & $(9.80\pm 4.75) \times 10^4$ &\centering $0.90 \pm 0.05$ \tabularnewline
Policy-Reuse  & $(5.57 \pm 2.59) \times 10^5$ &\centering $0.92 \pm 0.04$ \tabularnewline
\midrule
\multicolumn{3}{c}{\textbf{Fire Crafting Table (FCT) - Hard}}\tabularnewline
\midrule

RAPid-KGE-UCB  & $(8.61 \pm 4.73)\times10^3$ &\centering $0.98 \pm 0.00$\tabularnewline
\rowcolor{gainsboro}
RAPid-KGE-UAB  & $(4.40 \pm 1.96)\times10^3$ &\centering $0.99 \pm 0.00$\tabularnewline
RAPid-EG  & $(5.12 \pm 2.84)\times10^3$  &\centering $0.98 \pm 0.00$\tabularnewline
Actor-Critic Transfer  & $(16.7 \pm 3.99) \times 10^4$ &\centering $0.90 \pm 0.04$ \tabularnewline
Policy-Reuse  & $(7.90 \pm 2.70) \times 10^5$ &\centering $0.94 \pm 0.03$ \tabularnewline
\midrule

\multicolumn{3}{c}{\textbf{Scrape Plank (SP)}}\tabularnewline
\midrule

RAPid-KGE-UCB  & $(6.82 \pm 3.83)\times10^3$ &\centering $0.86 \pm 0.03$\tabularnewline
RAPid-KGE-UAB  & $(2.01 \pm 1.01)\times10^3$ &\centering $0.87 \pm 0.03$\tabularnewline
\rowcolor{gainsboro}
RAPid-EG  & $(17.2 \pm 7.38)\times10^2$ &\centering $0.96 \pm 0.01$\tabularnewline
Actor-Critic Transfer  & $(6.86  \pm 3.89) \times 10^4 $&\centering $0.93 \pm 0.03$ \tabularnewline
Policy-Reuse  & $(19.3 \pm 9.76) \times 10^4 $ &\centering $0.94 \pm 0.03$\tabularnewline
\bottomrule
\end{tabular}
\caption{\small Results of each agent's performance on $5$ novel scenarios. In all the novelties, RAPid-Learn consistently outperforms baseline approaches. (best performing agent highlighted.)}
\label{tab:results}
\end{center}
\end{table}

\subsection{Results}
We evaluate $3$ versions of RAPid-Learn along with $2$ baselines on $5$ novel scenarios (shown in Table~\ref{tab:results}). Time to adapt is the number of time steps each agent takes to adapt to a novelty (policy convergence given by a pre-defined convergence criteria\footnote{Convergence criteria details in Appendix Section~\ref{sec:convergence_criteria}}. Post-novelty performance is the success rate achieved by evaluating each agent on $100$ episodes in $10$ independent trials. In each episode we set a budget on the number of time steps the agent is given to execute the learned executor (if the budget is used up, we report an unsuccessful episode and assign $0$ score to that episode). A score of $1$ is recorded for each successful episode. After running $10$ independent trials with different random seeds, we report the mean and standard deviations. 

Table~\ref{tab:results} shows that our method is better at adapting to the novel scenarios and time to adapt is significantly lower than the baselines. In some of the cases, the baselines do not even converge to find a solution.
Fig \ref{fig:plots} compares the learning curves of the three different types of \textit{knowledge-guided-exploration} approaches described in this paper. From the learning curves, it is evident that the number of interactions taken to adapt to the novelty is directly dependent on the difficulty of the novelty.
In more complex novelties such as rubber-tree hard, we show
\emph{Knowledge-guided-exploration} increases sample efficiency by directing exploration for the agent in these novelties, as evident from the learning curves. The transfer learning baselines on these particular novelties consistently take $2$ (Table~\ref{tab:results}) orders of magnitude more interactions to adapt to the novelty.



\subsubsection{Statistical Significance}\label{statistical_significance}

To demonstrate that the average success rate of RAPid-Learn is consistently higher than the baseline approaches, we perform an unpaired t-test~\cite{kim2015t}. For the experiment, we consider a confidence interval of $95\%$ and evaluate the \emph{p-value} between the best performing RAPid-Learn approach and the two transfer learning baseline approaches (Actor-critic transfer, Policy reuse). Table~\ref{stat_sig} shows the results of the unpaired t-test. Thus, through the results, we see that our proposed approach, RAPid-Learn has a consistent performance in the post-novelty success rate. The results are always statistically significant, except in the case of the \emph{Scrape Plank} novelty. In all the novel scenarios, RAPid-Learn has a much more sample efficient performance.  
Thus, RAPid-Learn not only achieves a better success rate, but also adapts to novel scenarios efficiently.

\begin{table}[h]
\footnotesize
\begin{center}
\begin{tabular}{>{\centering}p{4.2cm}>{\centering}p{1.2cm}>{\centering}p{1.8cm}p{0.1cm}}

\arrayrulecolor{taupegray}\toprule

\textbf{\begin{small}Methods\end{small}} & {\textbf{\begin{small}p-value\end{small} }} & \textbf{\begin{small}Statistical significance\end{small} }\tabularnewline
\midrule
\multicolumn{3}{c}{\textbf{Axe to Break (ATB) + Fire Crafting Table (FCT) - Easy}} \tabularnewline
\midrule
 

RAPid-KGE-UCB $\leftrightarrow$ Actor-critic & $0.0062$ &\centering Yes\tabularnewline
RAPid-KGE-UCB $\leftrightarrow$ Policy Reuse & $0.0012$ &\centering Yes\tabularnewline

\midrule
\multicolumn{3}{c}{\textbf{Axe To Break (ATB) - Hard}} \tabularnewline
\midrule
 

RAPid-KGE-UCB $\leftrightarrow$ Actor-critic & $0.00336$ &\centering Yes\tabularnewline
RAPid-KGE-UCB $\leftrightarrow$ Policy Reuse & $0.0012$ &\centering Yes\tabularnewline

\midrule
\multicolumn{3}{c}{\textbf{Fire Crafting Table (FCT) - Hard}}\tabularnewline
\midrule

RAPid-KGE-UAB $\leftrightarrow$ Actor-critic & $< 0.0001$ &\centering Yes\tabularnewline

RAPid-KGE-UAB $\leftrightarrow$ Policy Reuse & $< 0.0001$ &\centering Yes\tabularnewline

\midrule

\multicolumn{3}{c}{\textbf{Scrape Plank (SP)}}\tabularnewline
\midrule

RAPid-KGE-EG $\leftrightarrow$ Actor-critic & $0.0077$ &\centering Yes\tabularnewline
RAPid-KGE-UCB $\leftrightarrow$ Policy Reuse & $0.0608$ &\centering Yes\tabularnewline

\bottomrule
\end{tabular}
\caption{\small Results of unpaired t-test between best performing RAPid-Learn and transfer learning baselines}\label{stat_sig}
\end{center}
\end{table}

\subsection{Discussion} 
\label{discussion}
\subsubsection{Dedicated Operator Failure}
In these set of experiments, we aim to find: \emph{Can the agent learn to succeed when an operator failure occurs due to one novelty introduced by the environment?} The novel scenarios corresponding to this are \emph{axe-to-break hard}, \emph{fire-crafting-table hard}, and \emph{scrape-plank}\footnote{Further details regarding the implemented novelties in Appendix Section~\ref{app:novelty_desc} } (see Table~\ref{tab:novelties}). The results in Table~\ref{tab:results} and the learning curves in~\ref{fig:plots}(c),~\ref{fig:plots}(d) show that RAPid-KGE-UCB and RAPid-KGE-UAB converge to a post-novelty performance of $96\%$, beating the baselines by $2$ orders of magnitude. The RAPid-Learn agent achieves a high success rate in the post-novelty performance in all the novel scenarios.
\subsubsection{Explicit Multiple Operator Failures}
In these set of experiments, we aim to find out: \emph{Can the agent learn to succeed when multiple independent operator failures occur due to multiple novelties introduced by the environment?} The environment introduces two novelties simultaneously, namely, \emph{fire-crafting-table easy} and \emph{axe-to-break easy} (\emph{ATB+FCT-Easy}). The effects of these novelties are independent in nature, i.e. the agent's performance over the \emph{fire-crafting-table easy} novelty does not depend on its performance of solving the \emph{axe-to-break easy} novelty. Thus, the agent learns two independent executors, one for each novelty.
The results in Table~\ref{tab:results} that RAPid-learn performs better than the baselines, achieving a post-novelty success performance of $97\%$.  Fig ~\ref{fig:plots}(b) compares different RAPid-Learn approaches.



\subsubsection{Implicit Multiple Operator Failures}
In these set of experiments, we aim to find: \emph{Can the agent learn to succeed when its learned action executor breaks another operator in the plan?} When the agent learns to adapt the \emph{rubber-tree hard} novelty by extracting \emph{rubber} in its inventory, it fails in executing the next operator in its original plan (it can no longer break the novel unbreakable \emph{rubber-tree}), running into another impasse. The agent can complete the task after it adapts to the multiple implicit novelties in the environment. Table~\ref{tab:results} shows that the baselines don't even converge, whereas RAPid-Learn adapts in about $28000$ time steps. Fig~\ref{fig:plots}(a) compares different RAPid-Learn approaches.

In all the novel scenarios in our experiments, the agent cannot succeed at the impasse by exploring the known predicates and can only succeed when it explores the environment at a sub-symbolic level. For example, in the \emph{rubber-tree-hard} novelty the agent's domain knowledge does not have any predicate for standing one block away from the \emph{tree} entity for performing the \emph{place} operator. Sub-symbolic learning materializes this notion through its sequence of actions in its policy, and in turn, in its action executor.

\section{Conclusion \& Future Work}
\label{sec:discussion_and_future_work}
We proposed a novel hybrid planning-deep RL based approach for open-world tasks which handles unanticipated changes to the task environment on-the-fly. Our proposed method utilizes domain knowledge to perform knowledge-guided-exploration in the novel environment and efficiently learns a novelty-handling policy mapped onto an action executor.  A rigorous evaluation of our domain-independent method in five novel scenarios demonstrated the significant performance improvement compared to state-of-the-art transfer learning approaches. We show novelty accommodation even in scenarios where baselines fail to converge. 

While our current implementation only handles cases where novelties lead to plan execution failure, specifically, execution impasses, we are working on extensions that can also handle other types of open-world novelties (beneficial, detrimental, irrelevant). Moreover, our approach assumes that a plan always exists, and a future direction would be one in which the agent needs to find a solution when the planning itself fails. Furthermore, while we have demonstrated the approach in a fully observable, deterministic environment, nothing hinges on these assumptions and thus this approach can be generalized to incorporate stochasticity and partial observability. Another limitation of our approach is that it assumes a fixed goal state. Additionally, we are working on extensions to abstract the learned executor to a symbolic operator, which is non-trivial when using function approximators. Finally, we would like to extend this approach to real-world robotic settings, where the cost of interactions is significantly higher.  

\section*{Acknowledgments}
This work was in part funded by DARPA grant W911NF-20-2-0006. We thank Gyan Tatiya for his contribution in developing NovelGridworlds and Lukas Weidenholdzer for his contribution in implementing the baselines.
\bibliographystyle{IEEEtran}
\balance
\bibliography{references}
\newpage
\appendix
\subsection{Novelty descriptions}
\label{app:novelty_desc}
In this section we describe the details of each novelty.
\subsubsection{Rubber tree}
In this novelty, as described in Table~$2$ the agent cannot extract \emph{rubber} from the regular trees anymore, but a new entity \emph{rubber-tree} appears in the environment and the agent needs to use the \emph{rubber-tree} for \emph{rubber} extraction.
We have two versions of this novelty, in the \emph{easy} version, the agent needs to select the \emph{tree-tap} and standing in front of a \emph{rubber-tree} it needs to use the action \emph{extract-rubber} to extract rubber. In the hard version, the agent's action space is augmented with a new action called \emph{place-tree-tap}, and in order to get rubber, the agent now needs to place the \emph{tree-tap} in front of the \emph{rubber-tree} and \emph{rubber} can only be extracted when the agent is in front of the placed \emph{tree-tap}. The agent's knowledge base does not have any \emph{in front of} predicate. The agent's executor learns the policy to place the new entity of \emph{tree-tap}, locate itself in front of it, and then extract \emph{rubber}. We observe that the agent is successful in synthesizing executors with predicates missing from the knowledge base.

\subsubsection{Axe to break}
\label{sec:axe_to_break}
In this novelty, a novel entity \emph{axe} appears in the environment and the agent cannot break \emph{trees} without holding the \emph{axe}. We implement two versions of the novelty, easy and hard. In the easy version, the \emph{axe} is present in the agent's inventory, and the agent needs to learn to select the \emph{axe} and use it in front of the \emph{tree} to get \emph{tree-log} into its inventory. In the hard version, the \emph{axe} is present in the environment, and the agent needs to find it to get into its inventory and then use it to break \emph{trees}. We observe that the agent is able to generate executors that can reason about the affordances of new entities in the environment, i.e. learn to pick up the \emph{axe}, select it, and then \emph{break} the \emph{tree} by locating itself next to the \emph{tree}.

\subsubsection{Fire crafting table}
\label{sec:fire_crafting_table}
In this novelty, the agent cannot access the \emph{crafting-table} since its set on \emph{fire}. Two novel entities \emph{water} and \emph{fire} appears in the environment, and novel actions \emph{spray} and \emph{select-water} are added to the list of available actions. In order to access the \emph{crafting-table} the agent needs to select \emph{water} and use the action \emph{spray} in front of the \emph{crafting-table} in order to remove it from \emph{fire}, and hence successfully access it. In the easy version, the \emph{water} is already present in the agent's inventory and in the hard version \emph{water} is present in the environment and the agent needs to collect it to use it.
The agent's sub-symbolic representation has information about the status of the \emph{fire} (on/off). The agent's learned executor in successfully dealing with more than one novel entities, and also novel predicates that are absent in the agent's knowledge base. 

\subsubsection{Scrape plank}
\label{sec:scrape_plank}
In this novelty, the \emph{break} action has no effect, and a novel action \emph{scrape-plank} is augmented in the agent's action space. The dynamics of the world changes in a way that the agent can no longer break \emph{trees} to get \emph{tree-logs}, but it can scrape planks from trees and using the action \emph{scrape-plank} in front of a \emph{tree} it gets $4$ \emph{planks} into its inventory. This novelty emphasizes on failure of effects of the operator rather than failure of the action itself. The executor learns a policy to generate \emph{planks}, by utilizing the \emph{scrape-plank} action to move ahead in the plan. Thus, the agent learns to tackle a novel scenario even when there are no new entities in the environment. 
\begin{algorithm}[b]
\small
\caption{\textit{\textbf{PlannableStateGenerator}} ($\Sigma, \mathcal{P}, o_i) \rightarrow \mathcal{S}_r$}
\label{alg:rewardfuncgen}
\begin{algorithmic}[1] 
\STATE $\mathcal{P}=\left[o_{1}, \ldots o_{\mid \mathcal{P}]}\right]$: Plan as an ordered list of operators 
\STATE $\Psi = \{{\psi_{o_1}, \psi_{o_2}, \ldots, \psi_{o_{\mid \mathcal{P}]}} }\}$: Set of preconditions of all the known operators
\STATE $\Omega = \{{\omega_{o_1}, \omega_{o_2}, \ldots, \omega_{o_{\mid \mathcal{P}]}} }\}$: Set of effects of all the known operators

\STATE $\mathcal{S}_r \rightarrow \emptyset$
\FOR{$o_j$ in \textit{reversed}$(\mathcal{P})$}
\IF{  {$o_j \neq o_i$} $\And$ $\psi_{o_j}  \not\supseteq \omega_{o_i}$}
\STATE$\mathcal{S}_r \cup \psi_{o_j}$
\FOR{$\omega \in \omega_{o_j}$ }
\IF{$\omega \subseteq \mathcal{S}_r$}
\STATE $\mathcal{S}_r \setminus \{\omega\}$
\ENDIF
\ENDFOR
\ENDIF
\STATE $\mathcal{S}_r \cup \omega_{o_i}$
\ENDFOR
\if
\RETURN $\mathcal{S}_r$
\end{algorithmic}
\end{algorithm}

\subsection{Additional Algorithms}
\label{sec:additional_algo}
We describe the algorithm which generates the set of plannable states. 

The agent accumulates the set of preconditions $\Psi$ and effects $\Omega$ of all known operators, using the  pre-novelty domain knowledge. Using that, it generates a set of plannable states $\mathcal{S}_r$, which contains: 1) the states that satisfy the effects  $\omega_{o_i}$ of the failed operator $o_i$ and 2) the states that satisfy the effects of all subsequent operators $\hat{o}\in\hat{\mathcal{O}}$ in the plan $\mathcal{P}$ (($d(\tilde{s}) \supseteq \omega_{\hat{o}}) \forall \hat{o} \in \hat{\mathcal{O}} $), where the preconditions of $\hat{o}$ contain the effects of $o$ ($\psi_{\hat{o}} \supseteq \omega_o$)(as shown in Algorithm~\ref{alg:rewardfuncgen}). This way we generate our algorithm generates a list of all the states from where the agent can successfully plan to reach the goal state. We provide this as a disjunction of condition 1 and 2 for the agent to satisfy at each time step. In other words, if the agent satisfies either 1 or 2, it gets a positive reward $\phi_1$ and the episode terminates. 



\subsection{Baseline Approaches}
\begin{figure*}[t]
	\centering
	\begin{minipage}{0.49\columnwidth}
		\centering
		\includegraphics[width=\textwidth]{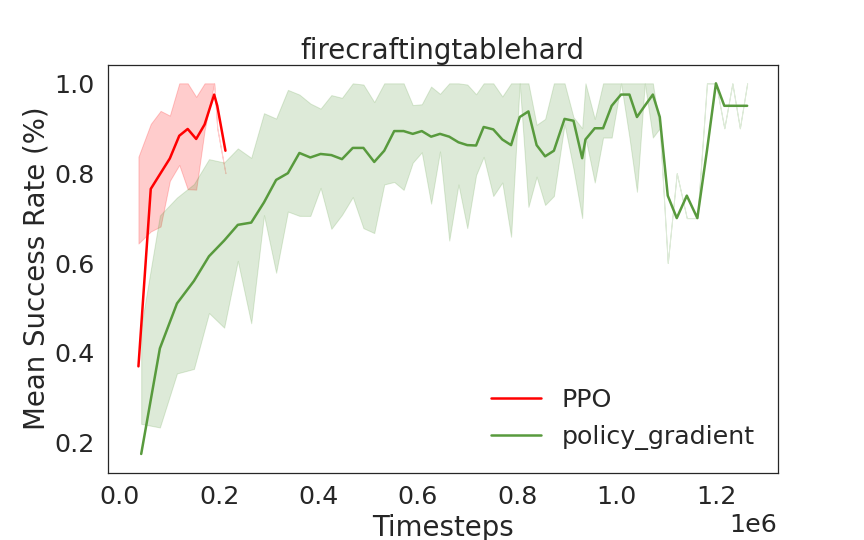}
		\small (a) Fire-crafting-table hard
		\label{fig:firecraftingtablehard}
	\end{minipage}%
	\begin{minipage}{0.49\columnwidth}
		\centering
		\includegraphics[width=\textwidth]{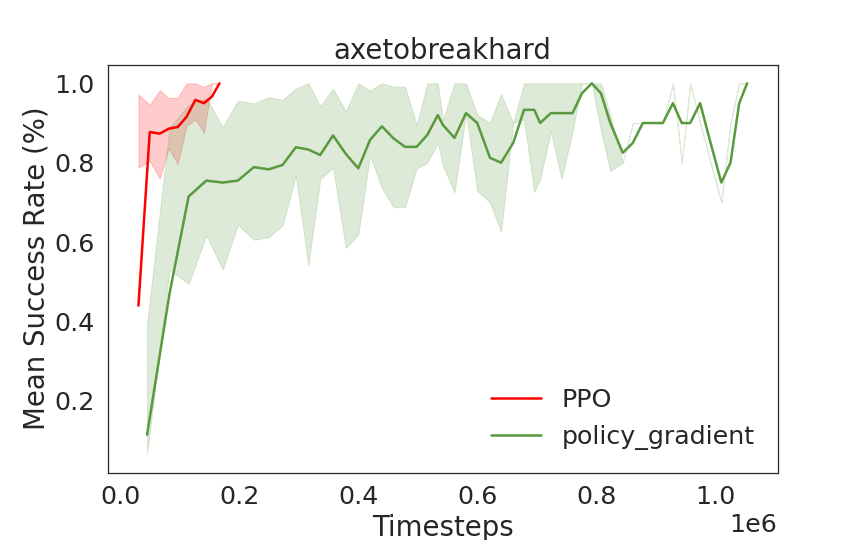}
        \small (b) Axe-to-break hard
		\label{fig:axetobreakhard}
    	\end{minipage}
	\begin{minipage}{0.49\columnwidth}
		\centering
		\includegraphics[width=\textwidth]{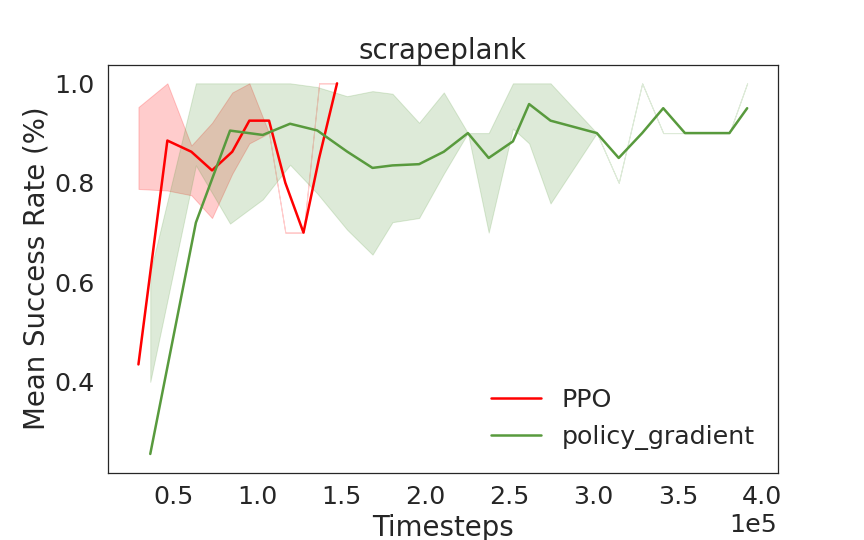}
        \small (c) Scrape plank
		\label{fig:rubber_tree_easy}
	\end{minipage}
	\begin{minipage}{0.49\columnwidth}
		\centering
		\includegraphics[width=\textwidth]{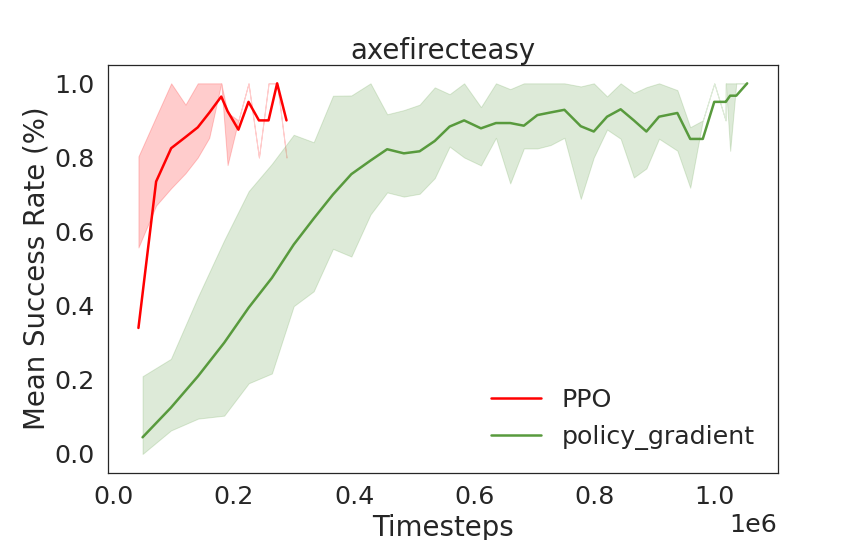}
        \small (d) Axetobreak+FCT(easy)
		\label{fig:axeFCT}
	\end{minipage}
	\caption{Baseline learning curves on the four novelties. Green curve shows the actor-critic transfer, and the red curve shows policy-reuse. Each plot demonstrates the performance during the novelty adaptation phase.}
    \label{fig:baselines_plots}
\end{figure*}

In this section we describe the details of the baselines, specifically, how they were trained for pre-novelty scenarios in Section~\ref{sec:baselines_prenovelty_training}, how they preformed in the pre-novelty scenario in Section~\ref{sec:baselines_prenovelty_performance}, followed by how transfer was performed upon novelty injections in Section~\ref{sec:baselines_transfer_approach}. Finally, Figure~\ref{fig:baselines_plots} compares the learning curves of these baselines on some novelties.

\subsubsection{Pre-novelty reward shaping}
\label{sec:baselines_prenovelty_training}
In early experiments, the pre-novelty task with sparse rewards proved intractable for pure RL approaches. Moreover, the RAPid-Learn agent has access to the full PDDL description of the environment, which heavily biases any comparison in its favor. We therefore employ strong reward shaping to allow the RL baseline to learn the prenovelty task. A small reward ($+50$) is given for completing intermediate steps (e.g. crafting sticks or planks) when they are needed for the next major step in the crafting process. Larger rewards are given for the subgoals of crafting the treetap ($+200$), extracting rubber ($+300$) and finally crafting the pogostick ($+1000$). Note that the rewards are only given if the intermediate step is the appropriate thing to do next. I.e. no reward is given for breaking a tree, if the agent already has a treetap and enough sticks and planks in its inventory to craft the pogostick. In this case, a reward would only be given for obtaining the missing ingredient (rubber).
Making the reward conditional on the stage within the crafting process was essential in getting the agent to learn the pre-novelty task. In early versions of the reward shaping, we attempted to give reward for completing intermediate steps (approaching a tree, crafting sticks, etc.). This was exploited by the agent, which learned that the reward could be optimized by periodically turning away and re-approaching a tree until the end of its life.

\subsubsection{Prenovelty performance}
\label{sec:baselines_prenovelty_performance}
The baselines as described~\ref{sec:baselines_prenovelty_training} were trained using a shaped reward. We evaluate their performance to make sure that they are at par with the RAPid-Learn approach (RAPid-Learn, being a planning agent, shows a $100\%$ performance in the pre-novelty scenario). Table~\ref{tab:baseline_prenovelty_performance} shows how many timesteps the pre-novelty models have experienced until reaching convergence and their performance on $100$ evaluation instances of the environment on $10$ random seeds. Results in Table~\ref{tab:baseline_prenovelty_performance} show that both the baseline agents converge to a policy which performs with a success of the task being achieved $95\%$ of the time.

\begin{table}[b]
\begin{center}
\small
\begin{tabular}{>{\centering}p{2.5cm}>{\centering}p{2.0cm}>{\centering}p{2.8cm}p{1.3cm}}
\arrayrulecolor{taupegray}\toprule
 \multicolumn{3}{c}{\textbf{Pre-novelty}} \tabularnewline
 \midrule
\centering \textbf{Agent} & \centering {\textbf{Time to learn\\(timesteps)}} &\centering \textbf{Pre-novelty performance\\(success rate \%)}\tabularnewline
 \midrule
 &\begin{small} \end{small}
 & \begin{small}Mean $\pm$ SD\end{small}
\tabularnewline
 \midrule
Actor-Critic Transfer  & $2.47 \times 10^5$ &\centering $0.95 \pm 0.22$ \tabularnewline
Policy Reuse  & $1.48 \times 10^6$ &\centering $0.95 \pm 0.22$ \tabularnewline
\bottomrule
\end{tabular}
\caption{Performance and timesteps trained for the best pre-novelty model that was later used in transfer learning the novelty environments}
\label{tab:baseline_prenovelty_performance}
\end{center}
\end{table}

\subsection{Baselines transfer mechanism}
\label{sec:baselines_transfer_approach}
The injected novelties alter the shape of the observation and action spaces of the environment. E.g.  the Fire crafting table novelty adds new observations for the \emph{water} item in the inventory and the LiDAR-sensor, as well as the \emph{select water} and \emph{spray} actions. 
In order to allow transferring the model from pre-novelty to post-novelty environments, we pad the observation and action spaces of all environments (pre- and post-novelty) with placeholder elements such that all of them have the same shape. Observation placeholders simply always return zero. When the model chooses to perform a placeholder action, the action actually passed on to the environment is always the \emph{approach treelog}.

\begin{figure*}[t]
	\centering
	\begin{minipage}{.49\columnwidth}
		\centering
		\includegraphics[width=\textwidth]{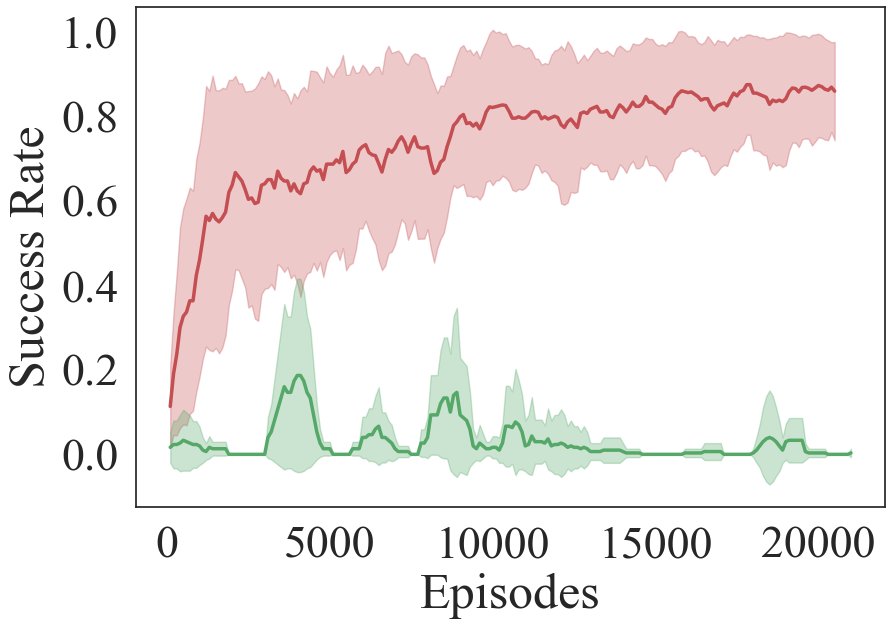}
        \small (a) FireCraftingTable Hard
		\label{fig:firecraftingtablehard}
	\end{minipage}%
	\begin{minipage}{0.49\columnwidth}
		\centering
		\includegraphics[width=\textwidth]{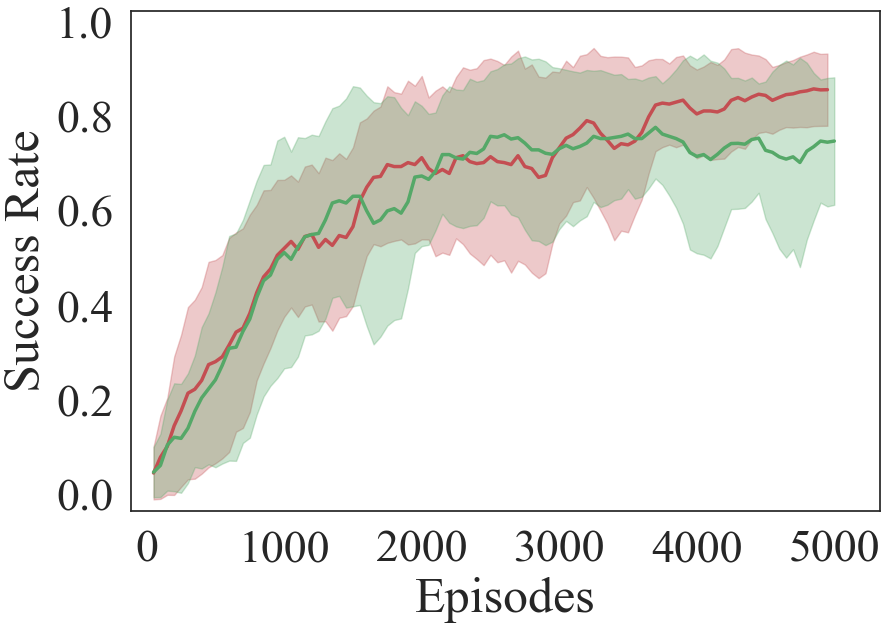}
        \small (b) Axe-to-break Hard
		\label{fig:axetobreakhard}
    	\end{minipage}
	\begin{minipage}{0.49\columnwidth}
		\centering
		\includegraphics[width=\textwidth]{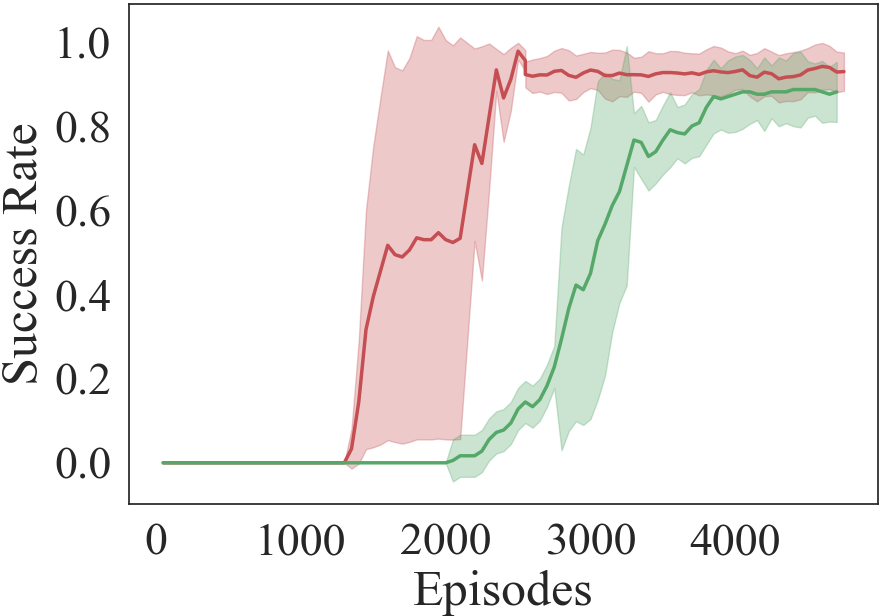}
        \small (c) Rubber-Tree Easy
		\label{fig:rubber_tree_easy}
	\end{minipage}
	\begin{minipage}{0.49\columnwidth}
		\centering
		\includegraphics[width=\textwidth]{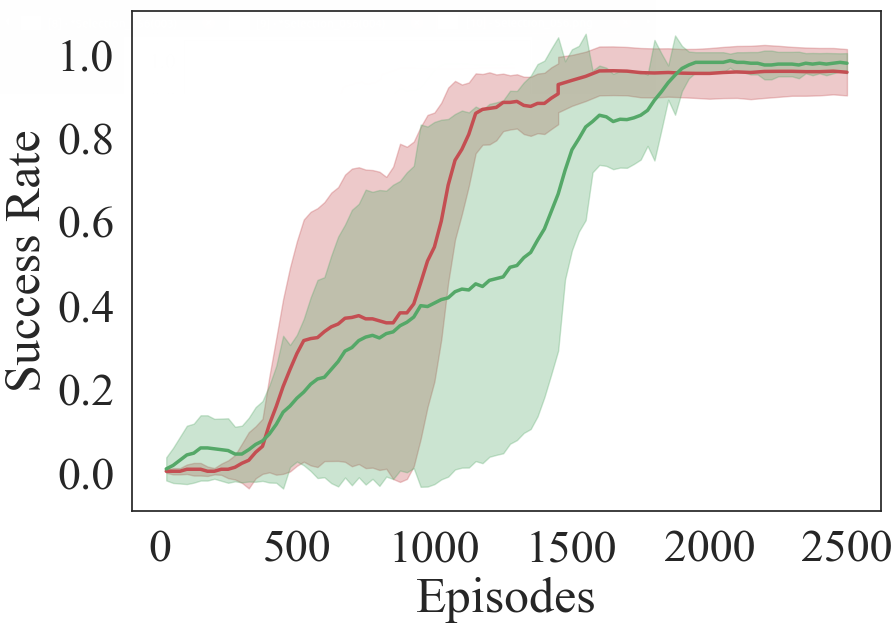}
        \small (d) Axe-to-break+FCT(easy)
		\label{fig:AxeFCT}
	\end{minipage}
	
	\begin{minipage}{.95\columnwidth}
	\vspace{1.25em}
		\includegraphics[width=1\textwidth, height=0.085\textwidth]{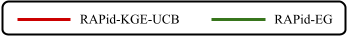}
		\label{fig:graph_label}
	\end{minipage}	
	\caption{The plots show the learning curves for four novelties on ablation study (no hierarchical-action operators were used). Red curve shows RAPid-Learn with knowledge-guided-exploration function, and green curve shows without the knowledge-guided-exploration but a vanilla epsilon-greedy exploration.}
    \label{fig:non_hier}
\end{figure*} 

\subsection{Ablation study}
To examine the impact of hierarchical actions combined with the knowledge-guided-exploration function, we perform experimental evaluation in absence of hierarchical actions. For our experiments, hierarchical actions are defined by the ({\tt approach}) operator, followed by the entity to approach. This hierarchical action is composed of the primitive actions ({\tt turn left, turn right} and {\tt move forward}). 
An example of an hierarchical action is {\tt approach tree\_log}, where the agent determines the location of one of the {\tt tree\_log} in the environment and formulates a plan to reach the entity. This plan is computed using the $A^*$ algorithm~\cite{hart1968formal}.

Incorporation of hierarchical actions simplifies novelty adaptation, as the agent has the opportunity to accommodate the novelty in fewer interactions with the environment. Fig~\ref{fig:non_hier}. We can observe that for the complex \emph{fire-crafting-table hard} novelty, the \emph{RAPid-EG} approach fails to converge to a successful policy. The difficulty for the agent to successfully complete the task can be attributed to two reasons:
\begin{itemize}
    \item The agent has to follow a series of movement actions to first collect the \emph{water} in its inventory, and then to reach the \emph{crafting table} where it needs to spray the \emph{water} by holding it, in order to diffuse the fire.
    \item While it performs the above mentioned steps, it should avoid reaching the state from which it cannot plan again. This will happen if the agent crafts crafts extra sticks in its inventory, making it impossible to craft a pogo stick with the resources available in the environment and in its inventory.
\end{itemize}
 
In absence of any informed exploration, the baseline approach 'RAPid-EG' takes longer to converge to a successful policy in absence of hierarchical actions.
Thus, we can see that hierarchical actions play a crucial role in increasing the sample efficiency of the outlined approach.
\subsection{Training details}
\subsubsection{Convergence criteria}
\label{sec:convergence_criteria}
To ensure that the learner has learned to reach the goal, we verify if the agent achieved the goal $\delta_g \! \geq \! \delta_G \!$ in the last $ \eta $ episodes with an average reward $\delta_r \! \geq \! \delta_R \!$. Furthermore, to ensure the agent has converged, and is not learning any further, we verify if the success rate for the agent in the past $\eta+\upsilon$ episodes is equal to the success rate for the agent in the past $\eta$ episodes.

\subsubsection{RAPid-Learn hyperparameters}
See Table~\ref{tab:hyper_rapid} for the hyperparameters used in the RAPid-Learn experiments.
\begin{table}[h]
\begin{center}
\small
\begin{tabular}{@{}cc@{}}
 \arrayrulecolor{taupegray}\toprule
    \textbf{Parameter} & \textbf{Value} \\ \midrule
    max-epsilon ($\epsilon_{max}$) & $0.3$ \\ \midrule
    min-epsilon ($\epsilon_{min}$) & $0.05$ \\ \midrule
    guided curriculum parameter ($\rho_{\text{max}}$) & $0.3$ \\ \midrule
    guided curriculum parameter ($\rho_{\text{min}}$) & $0.05$ \\ \midrule
    parameter decay speed & $\ln(0.01)/2000$ \\ \midrule
    UCB- parameter ($c$) & $0.0005$ \\ \midrule
    UAB-parameter( $\mu$) &$2$  \\ \midrule
    update\_rate & $10$ \\ \midrule
    max episodes ($e_{max}$) & $100000$ \\ \midrule
    max timesteps ($T$) & $300$ \\ \midrule
    Hidden Layers & $24$ (single layer network) \\ \midrule
    Discount factor ($\gamma$) & $0.98$ \\ \midrule
    learning rate ($\alpha$) & $1e-3$ \\ \midrule
    reward threshold ($\delta_{R}$) & $900$ \\ \midrule
    episodes threshold ($\delta_{G}$) & $100$ \\ \midrule
    no of success ($\eta$) & $96$ \\ \midrule
    positive reinforcement ($\phi_1$) & $1000$ \\ \midrule
    negative reinforcement ($\phi_2$) & $-350$ \\ \midrule
\end{tabular}
\end{center}
\caption{Hyperparameters for RAPid-Learn}
\label{tab:hyper_rapid}
\end{table}
\begin{table}[b]
  \small
  \begin{center}
\begin{tabular}{@{}cc@{}}
     \arrayrulecolor{taupegray} \toprule
    \textbf{Parameter} & \textbf{Value} \\  \midrule
    learning\_rate & $3e-4 $ \\ \midrule
    batch\_size & $64 $ \\ \midrule
    n\_epochs & $10 $ \\ \midrule
    n\_steps & $2048 $ \\ \midrule
    gamma & $0.98 $ \\ \midrule
    gae\_lambda & $0.95 $ \\ \midrule
    clip\_range & $0.2 $ \\ \midrule
    max\_grad\_norm & $0.5 $ \\ \midrule
    ent\_coef & $0.0 $ \\ \midrule
    vf\_coef & $0.5 $ \\ \midrule
    optimiser & Adam \\ \midrule
  \end{tabular}
  \end{center}
  \caption{Hyperparameters used for the PPO models (both pre- and post-novelty).}\label{tab:hyper_baselines}
\end{table}
\subsubsection{Baseline Hyperparameters}
See Table~\ref{tab:hyper_baselines} for an overview of the hyperparameters used to train the actor-critic transfer baselines. Note that with the exception of gamma, these are the default parameters used in \cite{stable-baselines3}. The Policy Reuse baseline models utilize the exact same implementation that underlies the Learner in RAPid-Learn, with identical hyperparameter settings, as shown in Table~\ref{tab:hyper_rapid}

\clearpage

\subsection{Domain PDDL}
Below is the original PDDL domain given to RAPid-Learn for the pogostick-crafting task inspired by Minecraft in novelgridworlds.
\begin{verbatim}
(:requirements :typing :strips :fluents)
(:types
    wall - physobj
    entity - physobj
    plank - physobj
    crafting_table - physobj
    tree_tap - physobj
    var - object
    rubber - physobj
    physobj - physical
    actor - physical
    pogo_stick - physobj
    stick - physobj
    tree_log - physobj
    air - physobj
)

(:predicates
    (holding ?v0 - physobj)
    (floating ?v0 - physobj)
    (facing ?v0 - physobj)
)

(:functions
    (world ?v0 - object)
    (inventory ?v0 - object)
)

(:action approach
    :parameters    (?physobj01 - physobj ?physobj02 - physobj )
    :precondition  (and
        (>= ( world ?physobj02) 1)
        (facing ?physobj01)
    )
    :effect  (and
        (facing ?physobj02)
        (not (facing ?physobj01))
    )
)

(:action crafttree_tap ; 
    :parameters    ()
    :precondition  (and
        (>= ( inventory plank) 4)
        (>= ( inventory stick) 1)
        (facing crafting_table)
    )
    :effect  (and
        (increase ( inventory tree_tap) 1)
        (decrease ( inventory plank) 5)
        (decrease ( inventory stick) 1)
    )
)

(:action craftplank
    :parameters    ()
    :precondition  (>= ( inventory tree_log) 1)
    :effect  (and
        (increase ( inventory plank) 4)
        (decrease ( inventory tree_log) 1)
    )
)

(:action break
    :parameters    ()
    :precondition  (and
        (facing tree_log)
        (not (floating tree_log))
    )
    :effect  (and
        (facing air)
        (not (facing tree_log))
        (increase ( inventory tree_log) 1)
        (increase ( world air) 1)
        (decrease ( world tree_log) 1)
    )
)

(:action craftstick
    :parameters    ()
    :precondition  (>= ( inventory plank) 2)
    :effect  (and
        (increase ( inventory stick) 4)
        (decrease ( inventory plank) 2)
    )
)

(:action extractrubber
    :parameters    ()
    :precondition  (and
        (>= ( inventory tree_tap) 1)
        (facing tree_log)
        (holding tree_tap)
    )
    :effect  (and
        (increase ( inventory rubber) 1)
    )
)

(:action craftpogo_stick
    :parameters    ()
    :precondition  (and
        (>= ( inventory plank) 2)
        (>= ( inventory stick) 4)
        (>= ( inventory rubber) 1)
        (facing crafting_table)
    )
    :effect  (and
        (increase ( inventory pogo_stick) 1)
        (decrease ( inventory plank) 2)
        (decrease ( inventory stick) 4)
        (decrease ( inventory rubber) 1)
    )
)

(:action select
    :parameters    (?physobj01 - physobj )
    :precondition  (and
        (>= ( inventory ?physobj01) 1)
        ; (holding air)
    )
    :effect  (and
        (holding ?physobj01)
        (not (holding air))
    )
)

)
\end{verbatim}

\end{document}